\documentclass[10pt,twocolumn,letterpaper]{article}

\usepackage{cvpr}
\usepackage{placeins} % To produce the REVIEW version
\usepackage[dvipsnames]{xcolor}

\newcommand{\PAR}[1]{\noindent{\bf #1~}}

\usepackage{comment}
\usepackage[acronym]{glossaries}
\usepackage{multirow}

\makeglossaries{}
\newacronym{sota}{SotA}{State-of-the-Art}
\newacronym{nn}{NN}{Nearest-Neighbor}
\newacronym{sfm}{SfM}{Structure-from-Motion}
\newacronym{tab}{Tab.}{Table}

\makeatletter
\DeclareRobustCommand\onedot{\futurelet\@let@token\@onedot}
\def\@onedot{\ifx\@let@token.\else.\null\fi\xspace}
\DeclareMathOperator*{\argmin}{arg\,min}

\def\eg{\emph{e.g}.} 
\def\ie{\emph{i.e}.} \def\Ie{\emph{I.e}.}
\def\cf{\emph{cf}.}

\def\etal{\emph{et al}.}
\makeatother

\definecolor{cvprblue}{rgb}{0.21,0.49,0.74}
\usepackage[pagebackref,breaklinks,colorlinks,citecolor=cvprblue]{hyperref}

\def\paperID{*****} % *** Enter the Paper ID here
\def\confName{3DV\xspace}
\def\confYear{2025\xspace}

\title{Obfuscation Based Privacy Preserving Representations are Recoverable Using Neighborhood Information}

\author{Kunal Chelani$^1$\thanks{Equal Contribution.}\\
\and
Assia Benbihi$^2$$^*$\\
\and
Fredrik Kahl$^1$\\
\and
Torsten Sattler$^2$
\and
Zuzana Kukelova$^3$
\and
$^1$Chalmers University of Technology\\
$^2$Czech Institute of Informatics, Robotics and Cybernetics, Czech Technical University in Prague\\
$^3$Visual Recognition Group, Faculty of Electrical Engineering, Czech Technical University in Prague \\
{\tt\small chelani@chalmers.se}
}

\begin{document}
\maketitle
\begin{abstract}
The rapid growth of AR/VR/MR applications and cloud-based visual localization has heightened concerns over user privacy. 
This privacy concern has been further escalated by the ability of deep neural networks to recover detailed images of a scene from a sparse set of 3D or 2D points and their descriptors -  the so-called inversion attacks.
Research on privacy-preserving localization has therefore focused on preventing such attacks through geometry obfuscation techniques like lifting points to higher dimensions or swapping coordinates. 
In this paper, we reveal a common vulnerability in these methods that allows approximate point recovery using known neighborhoods.
We further show that these neighborhoods can be computed by learning to identify descriptors that co-occur in neighborhoods.
Extensive experiments demonstrate that all existing geometric obfuscation schemes remain susceptible to such recovery, challenging their claims of being privacy-preserving.
  % Rapid growth in the popularity of AR/VR/MR applications and cloud-based visual localization systems has given rise to an increased focus on the privacy of user content in the localization process.
  % This privacy concern has been further escalated by the ability of deep neural networks to recover detailed images of a scene from a sparse set of 3D or 2D points and their descriptors -  the so-called inversion attacks.
  % Research on privacy-preserving localization has therefore focused on preventing these inversion attacks on both the query image keypoints and the 3D points of the scene map.
  % To this end, several geometry obfuscation techniques that lift points to higher-dimensional spaces, i.e., lines or planes, or that swap coordinates between points %
  % have been proposed. 
  % In this paper, we point to a common weakness of these obfuscations that allows to recover approximations of the original point positions under the assumption of known neighborhoods. 
  % We further show that these neighborhoods can be computed by learning to identify descriptors that co-occur in neighborhoods. 
  % Extensive experiments show that our approach for point recovery is practically applicable to all existing geometric obfuscation schemes. 
  % Our results show that these schemes should not be considered privacy-preserving, even though they are claimed to be privacy-preserving. 
  Code will be available at \url{https://github.com/kunalchelani/RecoverPointsNeighborhood}.
\end{abstract}

\section{Introduction}
Visual localization estimates the position and orientation of a camera in a given scene and is central for autonomous navigation~\cite{suomela2023benchmarking,thoma2019mapping}, Simultaneous Localization and Mapping (SLAM)~\cite{cummins2010fab,dissanayake2001solution}, 
Augmented and Virtual Reality (AR/VR)~\cite{guzov2021human,pons2023interaction,sarlin2022lamar}, and \gls{sfm}~\cite{schonberger2016structure,schonberger2016pixelwise}. 
The best performing methods represent the scene with a 3D map, \eg, a Structure-from-Motion (SfM) point cloud~\cite{sarlin2019coarse,Sattler2017PAMI}.
To localize a given query image, they match the 
descriptors of 2D local features~\cite{Lowe04IJCV,detone2018superpoint} extracted from the query image against the descriptors of the 3D points in the map. 
The resulting 2D-3D point correspondences are used for camera pose estimation~\cite{fischler1981random,hartley2003multiple,Kukelova13ICCV,Kukelova2016CVPR,Larsson2019ICCV}. 
Such feature-based approaches are known to handle challenging conditions and to provide accurate pose estimates~\cite{Toft2022TPAMI,sarlin2022lamar}. 
However, they also pose a potential privacy risk because of \textit{inversion attacks}~\cite{pittaluga2019revealing}: 
it is possible to recover the query image in high detail from the 2D image features with inversion networks~\cite{pittaluga2019revealing,dangwal2021analysis}.
One can also recover the map's content from the 3D points and their descriptors~\cite{pittaluga2019revealing,song2020PP}.
Thus, feature-based methods cannot be directly applied in settings where privacy is of concern~\cite{speciale2019privacy,speciale2019privacy2d}, %
such as when a user sends data to a localization service in the cloud or when 3D maps are stored on an external server. 

\begin{figure*}[t!]
    \centering
    \includegraphics[width=0.86\textwidth]{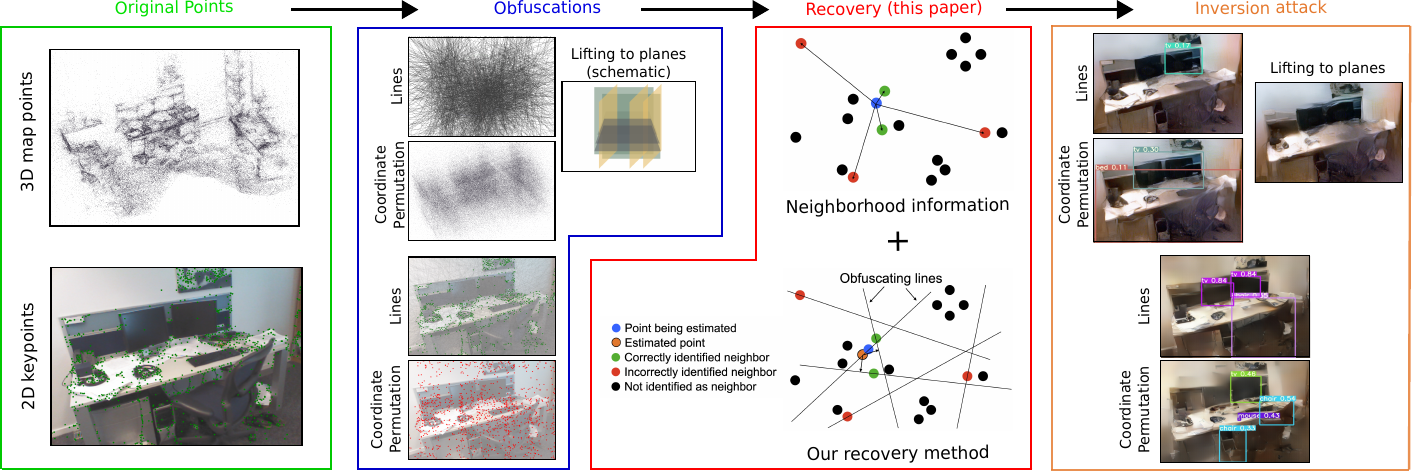}
    \caption{\small{\textbf{Geometry obfuscations allow the recovery of image details.} 
    The \textcolor{green}{original} point representations are privacy revealing as full images can be recovered from them~\cite{pittaluga2019revealing}.
    Different \textcolor{blue}{obfuscation schemes} are used to modify them.
    In \textcolor{red}{this paper}, we show that given neighborhood information, %
    it is possible to approximately recover the original %
    point positions, again enabling \textcolor{orange}{image recovery}.%
    }}
    \label{fig:teaser}
\end{figure*}

% Privacy-preserving localization methods aim to prevent such content recovery
% and mainly fall into two categories: 
% \textit{descriptor obfuscation} approaches aim to modify descriptors so as to not allow inversion while still allowing accurate 2D-3D matching~\cite{dusmanu2021privacy,ng2022ninjadesc,pittaluga2023iccv,pietrantoni2023segloc}. 
% \textit{Geometry obfuscation} approaches replace each 2D or 3D point with a potentially infinite set of points~\cite{speciale2019privacy,speciale2019privacy2d,geppert2020privacy,geppert2021privacy,geppert2022privacy,lee2023paired,pan2023privacy,shibuya2020privacy,moon2024efficient}.
Privacy-preserving localization methods aim to prevent content recovery and mainly fall into two categories: \textit{descriptor obfuscation} approaches, which modify descriptors to prevent inversion while enabling accurate 2D-3D matching~\cite{dusmanu2021privacy,ng2022ninjadesc,pittaluga2023iccv,pietrantoni2023segloc}, and \textit{geometry obfuscation} approaches, which replace each 2D or 3D point with a potentially infinite set of points~\cite{speciale2019privacy,speciale2019privacy2d,geppert2020privacy,geppert2021privacy,geppert2022privacy,lee2023paired,pan2023privacy,shibuya2020privacy,moon2024efficient}.
% For example, replacing a 3D point with a 3D line amounts to replacing the point with an infinite set of points forming the line passing through this point~\cite{speciale2019privacy,lee2023paired} (see~\cref{fig:teaser}).
An example of geometry obfuscation is lifting points to lines, which replaces each point with an infinite set of points lying on a line through the corresponding point~\cite{speciale2019privacy,lee2023paired} (see~\cref{fig:teaser}).
By substituting points with potentially infinite sets, these methods prevent the direct application of inversion attacks~\cite{pittaluga2019revealing,dusmanu2021privacy}. Geometric obfuscation approaches carefully design the function mapping a point to a set of points so that the resulting sets still enable pose estimation; for instance, in~\cite{speciale2019privacy}, 2D-3D point matches are replaced by 2D point-to-3D line matches.

Geometry obfuscation methods are considered privacy-preserving by the community since it is unclear how to recover the original point positions from the obfuscations.
However, none of the previous work proves that approximating the original point positions is impossible. 
On the contrary, \cite{chelani2021privacy} reveals the need for more scrutiny before claiming that a method is privacy-preserving: they show that when each 3D point is obfuscated by a line passing through the point with a random direction~\cite{speciale2019privacy}, it is possible to approximate the original 3D point position, thus enabling an inversion attack~\cite{pittaluga2019revealing}. 
Their approach is based on two key insights:
(1) the closest points on two such 3D lines are likely to be relatively close to the original 3D points. 
(2) to recover the position of a point $\mathbf{X}_i$ obfuscated by a line $\mathbf{l}_i$, it is important to know the neighbors of $\mathbf{l}_i$ (defined as the set of lines $\{\mathbf{l}_j\}$ corresponding to original 3D points that are neighbors of $\mathbf{X}_i$).
\cite{chelani2021privacy} cannot be applied to 2D point obfuscations~\cite{speciale2019privacy2d,pan2023privacy} and does not generalize since property (1) does not hold for all obfuscation schemes~\cite{speciale2019privacy2d,geppert2022privacy,pan2023privacy}.

Inspired by observation (2) from~\cite{chelani2021privacy}, we derive a novel and conceptually simple method for approximating the positions of the original points that is applicable for \textit{all} of the currently proposed geometry obfuscation schemes~\cite{speciale2019privacy,speciale2019privacy2d,lee2023paired,pan2023privacy,geppert2022privacy,moon2024efficient}. 
Our approach uses information about neighborhoods, \ie, about which obfuscated points correspond to nearest neighboring original points, and is computationally efficient\footnote{In some cases, \eg, ~\cite{lee2023paired}, our point recovery is faster to compute than obfuscating the points in the first place.}. %
% We show that the approximate point positions obtained with our method enable inversion attacks~\cite{pittaluga2019revealing,dusmanu2021privacy}. 
% Through extensive qualitative and quantitative experiments, we show that the proposed method is robust to errors in the neighborhoods, \ie, it is not necessary to have access to neighborhoods that contain only the nearest neighbors of the original points.
% We also demonstrate a simple way to learn neighborhoods from the descriptors associated with the geometric obfuscations, which need to be provided to enable visual localization. 
% Our approach for computing neighborhoods serves as a proof-of-concept to show that our scheme for recovering points from geometric obfuscations is practically applicable. 
We show that the approximate point positions obtained with our method enable inversion attacks~\cite{pittaluga2019revealing,dusmanu2021privacy}.
Through extensive qualitative and quantitative experiments, we demonstrate that the proposed method is robust to errors in neighborhoods, \ie, it does not require access to neighborhoods containing only the nearest neighbors of the original points. Additionally, we present a simple approach for learning neighborhoods from the descriptors associated with geometric obfuscations, which must be provided to enable visual localization.
Our method for computing neighborhoods serves as a proof of concept, demonstrating that our scheme for recovering points from geometric obfuscations is practically applicable.

In summary, this paper makes the following contributions: 
\textbf{(1)} we present a novel framework for recovering approximate point positions from obfuscated scene representations. 
Our framework relies on neighborhood information and is applicable to all geometry obfuscation schemes from the literature. 
\textbf{(2)} we propose a learning-based approach for computing the required neighborhoods from the descriptors used for visual localization. 
\textbf{(3)} extensive experiments with neighborhoods provided by an oracle and our approach show the effectiveness of our recovery framework. 
Our results show that methods that are currently considered privacy-preserving do not in fact guarantee privacy, and highlight the need to derive clear conditions under which privacy can be guaranteed when proposing privacy-preserving localization approaches.

\section{Related Work}
\PAR{Visual Localization.}
State-of-the-art localization methods rely on features and 2D-3D correspondences between query images and the map. 
These matches are fed into a robust estimation framework~\cite{fischler1981random,chum2003locally,barath2022space,barath2022learning} to estimate the pose of a camera~\cite{Li2012ECCV,Sattler2017PAMI,sarlin2019coarse,HumenbergerX20Kapture,Panek2022ECCV}.
The 3D scene map is generally represented as 3D points generated using Structure-from-Motion~\cite{schonberger2016pixelwise,schonberger2016structure} or SLAM pipelines~\cite{shibuya2020privacy}.
A drawback of feature-based methods is that sparse sets of points and descriptors are vulnerable to inversion attacks in which a neural network recovers detailed images of the scene from the points and their associated descriptors~\cite{pittaluga2019revealing,song2020PP,dangwal2021analysis,dusmanu2021privacy,pietrantoni2023segloc,mahendran2015understanding,dosovitskiy2016inverting,do2022learning,weinzaepfel2011reconstructing,vondrick2013hoggles,zeiler2014visualizing,yosinski2015understanding,kato2014image,dosovitskiy2016generating}.
While sparsifying the set of points improves privacy by reducing the inversion performance, it comes at the cost of reduced localization accuracy.

Alternative localization methods include Scene Coordinate Regression (SCR)~\cite{shotton2013scene,brachmann2021visual,brachmann2023accelerated} where the 3D map is represented by a neural network that predicts the 3D coordinates of every image pixel, resulting in 2D-3D correspondences.
SCR is said to be inherently privacy-preserving~\cite{Zhou2022ECCV} because there is no set of 2D or 3D points to run the inversion attack~\cite{pittaluga2019revealing,dusmanu2021privacy} on.
However, current SCR methods~\cite{brachmann2017dsac,brachmann2019neural,luo2020aslfeat} do not scale and do not handle challenging conditions as well as feature-based methods although these limitations are investigated~\cite{brachmann2023accelerated}.
Absolute Pose Regression (APR)~\cite{shavit2021learning,sattler2019understanding} and Relative Pose Regression (RPR)~\cite{Balntas2018RelocNetCM,abouelnaga2021distillpose} methods are end-to-end localization alternatives that share similar characteristics: they are inherently privacy-preserving but their performance falls behind feature-based methods, as pointed in~\cite{sattler2019understanding}.
In this paper, we analyze the privacy properties of features-based methods that remain the gold standard for accurate, robust, and efficient localization.

\PAR{Privacy Aspects of Visual Localization.}
Cloud-based localization services require the exchange of information about the scene between the client and the server, leaving several cracks for possible privacy leaks.
Naturally, such services should preserve the privacy of the 3D maps stored online from a curious/malicious server \cite{dangwal2021analysis,speciale2019privacy,pan2023privacy,geppert2022privacy,chelani2021privacy}.
They should also preserve the client's private information that is potentially sent to the server for localization, as part of the query image~\cite{speciale2019privacy2d,ng2022ninjadesc,dangwal2021analysis}.
As noted in~\cite{geppert2022privacy},
even the server knowing the client's accurate pose can potentially be a privacy risk.
It is also shown that even the minimal requirements for running a robust localization service - returning the camera pose to the client - enable the approximate recovery of the scene layout by a malicious third party\cite{Chelani_2023_CVPR}.
In this paper, we analyze the extent to which privacy-preserving geometric obfuscations can reveal private content from a 3D map stored on a server and from a client's query image.

\PAR{Privacy-Preserving Representations.}
Inversion attacks~\cite{pittaluga2019revealing,dusmanu2021privacy} take as input sparse feature maps to produce detailed images of the scene. 
The feature maps are made of descriptors located at keypoint positions and the keypoints are either 2D points or the projection of 3D points onto the image. %
Therefore, there are two obvious ways to counter such an attack by preventing the construction of the feature map: i) descriptor obfuscations that preserve the point information but modify the descriptors so that localization remains possible but not the inversion~\cite{dusmanu2021privacy,ng2022ninjadesc,pittaluga2023iccv},
ii) and geometric obfuscations that modify the geometry of the points. %

The first geometric methods obfuscate points with random lines~\cite{speciale2019privacy,speciale2019privacy2d,geppert2020privacy,geppert2022privacy} but \cite{chelani2021privacy} later shows that the 3D lines~\cite{speciale2019privacy} are not as privacy-preserving as originally claimed: the original points can be approximated using the geometry preserved in the random 3D lines.
\cite{chelani2021privacy} exploit the spatial distribution of the lines to estimate the points' nearest neighbors in the original space and then estimate the point positions that best agree with the neighborhood.
Subsequent works account for this important limitation to design geometric obfuscations that are less susceptible to recovery with~\cite{chelani2021privacy}.
One solution is to modify the distribution of line directions by lifting points to paired-point lines~\cite{lee2023paired} so that one line contains two points instead of one or constraining the lines to intersect at specific points~\cite{moon2024efficient}.
To further reduce the spatial correlation between the original points and their obfuscated representations, \cite{geppert2022privacy} lifts points to parallel planes and \cite{pan2023privacy} permutes coordinates of pairs of points, which prevents the estimation of nearest neighbors based on the geometric distances between obfuscations.
Overall, these methods obfuscate the position of the points while still allowing localization.
Here, we question their claim to be privacy-preserving: we reveal a weakness common to all the obfuscation schemes and propose a generic method that approximately recovers the original points from \textit{all} obfuscations when information on their neighborhood is available.

\section{Geometric Obfuscation of Points}
\label{sec:definition}

This section provides a general definition of obfuscations applied to points. 
Based on this definition, Sec.~\ref{sec:recovery} then proposes an attack that %
recovers approximate positions of the original points from an obfuscated representation given the knowledge about the neighborhoods of the original points.

\PAR{Definition: Geometry obfuscation.} A \textit{geometry obfuscation} applied to a point $x \in \mathbb{R}^m$ %
is a mapping
\begin{equation}\label{eq:obf_mapping}
\mathcal{O} : \mathbb{R}^m \rightarrow \mathcal{P}(\mathbb{R}^m)    \enspace ,
\end{equation}
where $\mathcal{P}(\mathbb{R}^m)$ is the power set of $\mathbb{R}^m$, \ie, the set of all subsets of $\mathbb{R}^m$. 
$\mathcal{O}$ maps a point in $\mathbb{R}^m$ to a (potentially infinite) set of points in $\mathbb{R}^m$.
Given a set of $n$ original points $P = \left\{ x_j \in \mathbb{R}^m,\; j =1,\dots, n \right\} $, an \emph{obfuscated representation} of $P$ is a set $\mathcal{O}(P) = \left\{ \mathcal{O}(x_j) \in  \mathcal{P}(\mathbb{R}^m),\; j =1,\dots, n \right\}$ obtained by obfuscating all of the $n$ points. 

This definition can be used to model all obfuscation schemes for 2D and 3D points in the literature. 
In the case of mapping a point $x_j$ to a line~\cite{speciale2019privacy,speciale2019privacy2d,lee2023paired,moon2024efficient} or a plane~\cite{geppert2022privacy}, $\mathcal{O}(x_j)$ contains all points on a line, respectively plane, that passes through $x_j$.\footnote{Note that while the set $\mathcal{O}(x_j)$ might be infinite, it can be represented compactly by the parameters of a line or plane.} 
In the case of obfuscation by coordinate permutation~\cite{pan2023privacy}, the set $\mathcal{O}(x_j)$ contains a single point $x'_j$ obtained by replacing one coordinate of $x_j$ with the corresponding coordinate of another point $x_i \in P \setminus x_j$.

\section{Recovering Obfuscated Points using Neighborhood Information}
\label{sec:recovery}
In this paper, we propose an attack designed to recover image content from obfuscated representations. 
It enables inversion attacks by approximating the original points from the obfuscated representations. 
Ideally, we would like to find the inverse of the obfuscation mapping $\mathcal{O}$ from~\eqref{eq:obf_mapping}, \ie,
\begin{equation}
\label{eq:inv_mapping}
\mathcal{O}^{-1} : \mathcal{P}(\mathbb{R}^m) \rightarrow  \mathbb{R}^m, \hspace{0.2cm} s.t \hspace{0.2cm} \mathcal{O}^{-1}(\mathcal{O}(x)) = x \enspace . 
\end{equation}
However, recovering such an inverse mapping is generally impossible~\footnote{For the coordinate permutation method~\cite{pan2023privacy}, the recovery of the inverse mapping is theoretically possible; however, it results in a combinatorial problem that can be computationally infeasible to solve~\cite{pan2023privacy}.}. 
Thus, we aim to %
find a mapping $\mathcal{R}$
\begin{equation}
\label{eq:R_mapping}
\mathcal{R} : \mathcal{P}(\mathbb{R}^m) \rightarrow  \mathbb{R}^m \enspace,   
\end{equation}
such that the set of points $\mathcal{R}(\mathcal{O}(P))=\left\{ \mathcal{R}(\mathcal{O}(x_j)) \in \mathbb{R}^m, \; j =1,\dots, n \right\}$ and their corresponding descriptors can reveal private information through inversion attacks~\cite{pittaluga2019revealing}. 
\Ie, the mapping $\mathcal{R}$ should facilitate recognizing objects, text, or persons in images recovered from the point positions in $\mathcal{R}(\mathcal{O}(P))$ and their descriptors.

Naturally, if the points $\mathcal{R}(\mathcal{O}(x_j))$ are close to the original points $x_j$ in the space $\mathbb{R}^m$, it can be expected that an inversion attack recovers a detailed image, potentially containing private information. %
Thus, if $d(\mathcal{R}(\mathcal{O}(x_j)),x_j)\leq \epsilon$, for some small $\epsilon$~\footnote{Here, $d$ is the Euclidean distance in $\mathbb{R}^m$.}, the obfuscation $\mathcal{O}$ cannot be considered as privacy preserving.
In this paper, we show that having information about the neighborhoods of the original points, we can compute a mapping $\mathcal{R}$ for which  $\epsilon$ 
is sufficiently small for most points. 
Thus, private details can be identified in images recovered from the $\mathcal{R}(\mathcal{O}(P))$.

\PAR{Recovering points using neighborhood information.}
Let us assume that for each obfuscated input point $\mathcal{O}(x_j)$ %
we are given a set of neighbors $\mathcal{N}(\mathcal{O}(x_j)) = \left\{\mathcal{O}(x_{j_i}) : j_i \in [j_1 \dots j_K]\right\}$. 
Here, $[j_1 \dots j_K]$ are the indices of the K nearest neighbors (in $\mathbb{R}^m$) of the point $x_j$ among all points in ${P}$. %
In other words, the set $\mathcal{N}(\mathcal{O}(x_j))$ contains the obfuscated representations corresponding to the K nearest neighbors in $P$ of the original point $x_j$. %
The assumption that each original point $x_j$ is contained in its set $\mathcal{O}(x_j)$, \ie,  $\forall j$ $x_j \in \mathcal{O}(x_j)$ holds for approaches that map $x$ to lines~\cite{speciale2019privacy,speciale2019privacy2d,lee2023paired,moon2024efficient} or planes~\cite{geppert2022privacy} passing through $x$. 
It does not hold for coordinate permutation-based obfuscation. 
However, as detailed below, for this case we can extend $\mathcal{O}(x_j)$ to include all points on $m$ lines passing through $\mathcal{O}(x_j)$ as one of them contains $x_j$.

We propose a strategy for computing a recovery mapping $\mathcal{R}$~\eqref{eq:R_mapping} based on the following fact: 
for a recovered point $\mathcal{R}(\mathcal{O}(x_j))$ for which $d(\mathcal{R}(\mathcal{O}(x_j)),x_j)\leq \epsilon$, for some small $\epsilon$, it holds that $d(\mathcal{R}(\mathcal{O}(x_j)),x_{j_i})\leq \epsilon_2$, for all K nearest neighbors $x_{j_i}, i = 1,\dots,K$ of $x_j$ and a small $\epsilon_2$~\footnote{Here $\epsilon_2 = \epsilon + d(x_j,x_{j_l})$ for the farthest neighbor $x_{j_l}$ from the K nearest neighbors of $x_j$.}.
Since by assumption $x_{j_i} \in \mathcal{O}(x_{j_i})$ for all $j_i$, it also holds that $d(\mathcal{R}(\mathcal{O}(x_j)),\mathcal{O}(x_{j_i})) \leq \epsilon_3$ for all $\mathcal{O}(x_{j_i}) \in \mathcal{N}(\mathcal{O}(x_j))$ and  $\epsilon_3 \leq \epsilon_2$, \ie, the recovered point $\mathcal{R}(\mathcal{O}(x_j))$ has a small distance from all obfuscated representations of the K nearest neighbors of the point $x_j$.
Since $x_j \in \mathcal{O}(x_j)$, we know that $\mathcal{O}(x_j)$ contains a point that is close to $x_j$ and also close to all $\mathcal{O}(x_{j_i}) \in \mathcal{N}(\mathcal{O}(x_j))$. 
We thus propose to compute a 
recovery mapping $\mathcal{R}$ %
by minimizing the cost function: 
\begin{equation}\label{eq:estimate}
    \mathcal{R}(\mathcal{O}(x_j)) = \argmin_{{x} \in \mathcal{O}(x_j)} \sum_{\mathcal{O}(x_{j_i}) \in \mathcal{N}(\mathcal{O}(x_j))} d(\mathcal{O}(x_{j_i}),{x}) \enspace.  %
\end{equation}
Here, $d(\mathcal{O}(x_{j_i}),x)$ is the  Euclidean distance of a point $x \in \mathbb{R}^m$ from its closest point in $\mathcal{O}(x_{j_i})$. 
Note that the point $\mathcal{R}(\mathcal{O}(x_j))$ that minimizes~\eqref{eq:estimate} can be far away from the true point position $x_j$. 
However, in our experience, using sufficiently many neighbors "pulls" $\mathcal{R}(\mathcal{O}(x_j))$ towards $x_j$. 
Also note that we solve %
\eqref{eq:estimate} per point $x_j$, rather than taking point estimates for the neighbors into account. 
This makes our recovery approach parallelizable. 

In the following, we concretely discuss how we compute the recovery mapping for individual obfuscation schemes.

\PAR{Points lifted to lines.} 
For the 
obfuscation that maps a point $x_{j}$ to a line passing through $x_j$, $\mathcal{O}(x_j)$ can be represented by the parameters of a line in $\mathbb{R}^m$. 
Thus, each point has a single degree of freedom, \ie, a shift along the line. 
We solve~\eqref{eq:estimate} via least-squares minimization in this variable, which can be easily implemented using existing optimization libraries, \eg,  Ceres~\cite{Agarwal_Ceres_Solver_2022}. 
In our experience, the choice of initialization for $\mathcal{R}(\mathcal{O}(x_j))$  is not critical (see the supp. mat. for details). 

In the case of paired-point lifting~\cite{lee2023paired}, each line passes through two original points. 
Each line contains the descriptors of both points, creating additional confusion as to which descriptor belongs to which point. 
Although this is not necessary for computing $\mathcal{R}$,
it is important for applying inversion attacks. %
We provide implementation details in Sec.~\ref{sec:evaluation}. 

In the case of 3D ray clouds~\cite{moon2024efficient}, each line passes through one original point and one of two additional center points.
The center points are derived by clustering the point cloud into two clusters which centers are the center points.
When solving~\eqref{eq:estimate}, we ignore all neighbors corresponding to lines passing through the same center as $\mathcal{O}(x_j)$.

\PAR{Points lifted to planes.} 
The method suggested in~\cite{geppert2022privacy} first splits the set of points into three disjoint sets $P_x$, $P_y$, and $P_z$. 
Each set is stored on a separate server. 
For the server storing $P_y$, each point $x \in P_y$ is represented by a plane parallel to the $\mathrm{xz}$-plane 
passing through the y-coordinate of $x$. 
Similar obfuscations are used for the points in $P_x$ and $P_z$~\cite{geppert2022privacy}. 
We consider the case where an attacker has access to all three servers - hence having three %
sets of parallel planes, each orthogonal to the other two.
This setting is realistic as access to all three servers is needed for 6D camera pose estimation. %

Each obfuscated point $\mathcal{O}(x)$ can be represented by two parameters corresponding to shifts along two basis vectors of a plane. 
Thus, each point has two degrees of freedom. %
We solve~\eqref{eq:estimate} via a %
two-variable optimization problem to find the position on a plane that minimizes the sum of distances to neighboring planes. 
As for line lifting, the initialization of the point positions is not critical (\cf{} supp. mat.).

\PAR{Coordinate permutation.} \label{method:cp} 
The obfuscation scheme based on permuting coordinates~\cite{pan2023privacy} randomly subdivides $P$ into pairs of points.
For a pair of points $x_j$ and $x_i$, \cite{pan2023privacy} %
randomly chooses a coordinate, \eg, the y-coordinate, %
and exchanges that coordinate between $x_j$ and $x_i$. %
The obfuscation thus maps a point $x_j$ to a single point $\mathcal{O}(x_j)$. %
Clearly, in general, it holds that $x_j \not= \mathcal{O}(x_j)$. 
However, note that $\mathcal{O}(x_j)$ shares $m-1$ coordinates with $x_j$. 
Thus $x_j$ is contained in one of $m$ lines, each parallel to one of the $m$ axes, passing through $\mathcal{O}(x_j)$~%
\cite{pan2023privacy}. 
These lines are used for camera pose estimation in~\cite{pan2023privacy}. 
We thus extend the obfuscation $\mathcal{O}(x_j)$ to contain all points on these lines, allowing us to use~\eqref{eq:estimate} to compute the mapping $\mathcal{R}$. 
In essence, this approach corresponds to lifting $x_j$ to $m$ lines, each one parallel to one of the $m$ coordinate axes.
In order to recover $\mathcal{R}(\mathcal{O}(x_j))$, we propose a method to determine along which of the $m$ lines $x_j$ has been moved (\ie, which coordinate of $x_j$ was exchanged). For details on this method, please see the supp. mat.
Given the line, we then use the same approach as for point-to-line lifting to compute $\mathcal{R}(\mathcal{O}(x_j))$.

\PAR{Robustness to imperfect neighborhoods.} 
So far, we have assumed that we are given an estimate of the neighborhood $\mathcal{N}(\mathcal{O}(x_j))$ for %
$\mathcal{O}(x_j)$. 
Sec.~\ref{sec:rec_nn} presents a practical approach for computing such estimates. 
However, the computed neighborhood estimates will contain outliers, \ie, obfuscated representations of points that do not correspond to one of the K nearest neighbors of $x_j$. 
As detailed above, we compute the mapping $\mathcal{R}$ via least-squares minimization, which is affected by outliers. 
To add robustness to outliers in the given neighborhood estimates, we include the minimization problem in a RANSAC-like loop~\cite{fischler1981random}.
In each iteration, %
we select a small number of neighbors and use them to compute an estimate for $\mathcal{R}(\mathcal{O}(x_j))$. %
We compute the distances of this estimate to all $\mathcal{O}(x_{j_i}) \in \mathcal{N}(\mathcal{O}(x_j))$ %
and classify $\mathcal{O}(x_{j_i}) \in \mathcal{N}(\mathcal{O}(x_j))$ into inliers and outliers using a %
pre-decided threshold $\delta$. %
We obtain the final estimate by solving \eqref{eq:estimate} %
over the largest %
inlier set found by this approach.
In practice, this approach is more robust than using a robust cost function in \eqref{eq:estimate}. %

\section{Estimating Neighborhoods From Descriptors}\label{sec:rec_nn}

Computing the recovered points $\mathcal{R}(\mathcal{O}(x_j))$ from the obfuscations $\mathcal{O}(x_j)$ using~\eqref{eq:estimate} assumes that we have information about the neighbors of each original point $x_j$. %
In ~\cite{chelani2021privacy}, such a neighborhood is geometrically estimated by using the distance between pairs of 3D lines as a proxy for the distance between the original points. 
However, their approach requires that the line directions are random and that lines are thus unlikely to intersect in 3D. 
This assumption does not hold for 2D lines, orthogonal 3D planes and ray clouds~\cite{moon2024efficient}.

In the context of visual localization, each obfuscated point is associated with a descriptor that is used for matching the 2D image query with the 3D map points.
We use these descriptors to estimate the required neighborhoods.

Intuitively, local structures (captured by the neighbors of a point) are not unique for each scene, but similar-looking structures can be found in other scenes. 
This motivates our learning-based approach for estimating the neighborhoods.  
Given enough scenes as training data, we let a neural network learn about such patterns, which in turn can be used to determine neighborhoods.
We pose the task of neighborhood estimation as a feature matching task~\cite{sarlin2020superglue,lindenberger2023lightglue,sun2021loftr}: given a set of descriptors, we learn a similarity score between all pairs of descriptors 
that is inversely proportional to the distance between the original points.
More specifically, the network takes as input the descriptors and outputs a row-normalized similarity matrix with high entries between the points that are likely to correspond to neighboring points.
The network is made of several self-attention blocks~\cite{vaswani2017attention} that draw contextual cues between the descriptors.
It is trained in a supervised manner with the binary cross-entropy loss.
The entry $(i,j)$ of the similarity matrix is positive if the $j^{\text{th}}$ point is within the $K$ closest points to the $i^{\text{th}}$ point.

\begin{table*}[t]
\fontsize{8}{6.2}\selectfont
\centering
\resizebox{0.6\textwidth}{!}{%
\begin{tabular}{l*{14}{c}}
\toprule
 & \multicolumn{6}{c}{\textbf{Lines}~\cite{speciale2019privacy2d}} & \multicolumn{6}{c}{\textbf{Coordinate Permutation}~\cite{pan2023privacy}}\\ 
\cmidrule(r){2-7} \cmidrule(r){8-13}
& \multicolumn{3}{c}{\textbf{7-scenes}~\cite{shotton2013scene}} & \multicolumn{3}{c}{\textbf{Cambridge}~\cite{kendall2015posenet}} & \multicolumn{3}{c}{\textbf{7-scenes}~\cite{shotton2013scene}} & \multicolumn{3}{c}{\textbf{Cambridge}~\cite{kendall2015posenet}}\\
\cmidrule(r){2-4} \cmidrule(r){5-7} \cmidrule(r){8-10} \cmidrule(r){11-13} 
In. & 5px & 10px & 25px & 5px & 10px & 25px & 5px & 10px & 25px & 5px & 10px & 25px\\ 
\cmidrule(r){2-4} \cmidrule(r){5-7} \cmidrule(r){8-10} \cmidrule(r){11-13} 
1.0 & 47.5 & 75.1 & 95.0 & 60.3 & 88.1 & 99.0 & 45.6 & 71.7 & 92.3 & 61.0 & 87.4 & 98.5\\ 
0.75 & 49.3 & 77.6 & 96.1 & 61.4 & 89.4 & 99.3 & 46.4 & 72.7 & 91.8 & 61.7 & 88.0 & 98.1\\ 
0.5 & 49.7 & 78.8 & 96.9 & 61.1 & 89.9 & 99.4 & 40.2 & 63.9 & 80.2 & 55.6 & 80.9 & 90.5\\ 
0.3 & 44.6 & 73.1 & 92.0 & 56.3 & 85.6 & 96.5 & 19.8 & 32.3 & 43.6 & 25.2 & 38.2 & 46.1\\ 
0.2 & 34.3 & 56.9 & 74.8 & 44.5 & 69.0 & 80.6 & 9.7 & 15.9 & 24.6 & 9.7 & 14.8 & 20.5\\ 
0.1 & 16.0 & 26.2 & 40.0 & 18.4 & 28.1 & 38.2 & 4.1 & 6.9 & 13.2 & 3.1 & 4.9 & 8.4\\ 
\bottomrule
\end{tabular}%
}
\caption{\textbf{Geometric accuracy of recovery from obfuscation of 2D SIFT~\cite{Lowe04IJCV} points.} Percentage of points recovered within error thresholds using oracle neighborhoods with different inlier ratios. Sizes: \textit{7-scenes}~\cite{shotton2013scene} - $640 \times 480$ and \textit{Cambridge}~\cite{kendall2015posenet} - $1024 \times 576$.}
\label{tab:geom_acc_2d}
\end{table*}

\begin{table*}[t]
\fontsize{8}{6.2}\selectfont
\centering
\resizebox{0.9\textwidth}{!}{%
\begin{tabular}{l*{16}{c}}
\toprule
 & \multicolumn{4}{c}{\textbf{PPL~\cite{lee2023paired}}} & \multicolumn{4}{c}{\textbf{Plane~\cite{geppert2022privacy}}} & \multicolumn{4}{c}{\textbf{CP~\cite{pan2023privacy}}} & \multicolumn{4}{c}{\textbf{Ray~\cite{moon2024efficient}}}\\ 
\cmidrule(r){2-5} \cmidrule(r){6-9} \cmidrule(r){10-13} \cmidrule(r){14-17}
 & \multicolumn{2}{c}{\textbf{7-scenes}~\cite{shotton2013scene}} & \multicolumn{2}{c}{\textbf{Cambridge}~\cite{kendall2015posenet}}
 & \multicolumn{2}{c}{\textbf{7-scenes}~\cite{shotton2013scene}} & \multicolumn{2}{c}{\textbf{Cambridge}~\cite{kendall2015posenet}}
 & \multicolumn{2}{c}{\textbf{7-scenes}~\cite{shotton2013scene}} & \multicolumn{2}{c}{\textbf{Cambridge}~\cite{kendall2015posenet}}
 & \multicolumn{2}{c}{\textbf{7-scenes}~\cite{shotton2013scene}} & \multicolumn{2}{c}{\textbf{Cambridge}~\cite{kendall2015posenet}}\\ 
\cmidrule(r){2-3} \cmidrule(r){4-5} \cmidrule(r){6-7} \cmidrule(r){8-9} \cmidrule(r){10-11} \cmidrule(r){12-13} \cmidrule(r){14-15} \cmidrule(r){16-17} 
In. & 10cm & 25cm & 25cm & 50cm &
      10cm & 25cm & 25cm & 50cm & 
      10cm & 25cm & 25cm & 50cm &
      10cm & 25cm & 25cm & 50cm\\ 
\cmidrule(r){2-3} \cmidrule(r){4-5} \cmidrule(r){6-7} \cmidrule(r){8-9} \cmidrule(r){10-11} \cmidrule(r){12-13} \cmidrule(r){14-15} \cmidrule(r){16-17}
1.0 & 94.6 & 97.3 & 69.2 & 83.2 & 93.4 & 97.5 & 65.2 & 81.1 & 88.2 & 94.5 & 65.3 & 81.0  &  94.6 & 97.9 & 72.1 & 83.6 \\ 
0.75 & 94.7 & 97.1 & 66.9 & 80.4 & 93.0 & 97.0 & 56.2 & 67.7 & 89.1 & 95.8 & 66.3 & 82.0 &  93.3 & 96.8 & 72.9 & 83.1 \\ 
0.5 & 95.0 & 97.2 & 67.2 & 79.3 & 82.8 & 88.7 & 33.2 & 38.5 & 67.7 & 75.0 & 61.4 & 72.5  &  91.9 & 95.7 & 74.4 & 84.1 \\ 
0.3  & 94.8 & 97.1 & 68.2 & 78.8 & 42.1 & 60.4 & 15.0 & 17.1 & 40.9 & 46.2 & 35.4 & 40.6 &  86.2 & 90.5 & 75.5 & 84.8 \\ 
0.2 & 94.0 & 96.8 & 69.0 & 78.4 & 20.9 & 39.6 & 8.1 & 9.4 & 31.1 & 35.1 & 24.1 & 27.2    &  78.7 & 83.6 & 75.0 & 84.2 \\ 
0.1 & 78.2 & 84.5 & 69.1 & 76.2 & 7.5 & 20.7 & 2.9 & 3.8 & 22.8 & 26.2 & 16.5 & 18.2     &  49.9 & 57.1 & 63.8 & 72.7\\ 
\bottomrule
\end{tabular}%
}
\caption{
\textbf{Geometric accuracy of recovery from obfuscation of 3D points (suing SIFT~\cite{Lowe04IJCV}).} Ratio of points recovered within error thresholds from oracle neighborhoods with different inlier ratios (In.) on 7-scenes~\cite{shotton2013scene} and Cambridge~\cite{kendall2015posenet} datasets.
The line obfuscations PPL~\cite{lee2023paired} and~\cite{moon2024efficient} are more susceptible to point recovery as compared to plane~\cite{geppert2022privacy} and point-permutation~\cite{pan2023privacy} obfuscations.
}
\label{tab:geom_acc_3d}
\end{table*}

% The network training is not tied to a dataset or to a specific point distribution but only to the type of local descriptor.
The network training is only tied to the descriptor, and thus, a network can be trained on any data where point positions and associated descriptors are available,
% and it can then be applied to the descriptors of the obfuscated representations.
As shown in the experiments, such a simple network predicts neighborhoods that are sufficiently reliable to allow the proposed recovery method to reveal private content robustly. 
Note that our approach is intended as a proof-of-concept to show that our attack from Sec.~\ref{sec:recovery} is practically feasible. 
We believe that better results can be obtained by tuning the network architecture and using larger training sets. 
However, such optimizations are outside the scope of this work.

\section{Experimental Evaluation}\label{sec:evaluation}

We evaluate the recovery method based on how well it recovers the points obfuscated by 6 different obfuscation schemes in 3D - random lines~\cite{speciale2019privacy} (OLC), two variants of paired points lines~\cite{lee2023paired} (PPL and PPL+), the default ray clouds~\cite{moon2024efficient}, planes~\cite{geppert2022privacy}, and Coordinate Permutation (CP)~\cite{pan2023privacy}.
In 2D, we evaluate the recovery from points obfuscated using random lines~\cite{speciale2019privacy2d} and with CP~\cite{pan2023privacy}.  
We experiment with two widely used descriptors: the hand-crafted SIFT~\cite{Lowe04IJCV} and the learning-based SuperPoint~\cite{detone2018superpoint}.
To visually assess the information revealed by the point recovery, 
we further invert the recovered points and their descriptors into images of the scene with an inversion network~\cite{pittaluga2019revealing,dusmanu2021privacy}.
For most of our analysis, we use oracle-provided neighborhoods, \ie, neighborhoods directly obtained from the nearest neighbors of the original points, rather than neighborhoods computed using our approach from Sec.~\ref{sec:rec_nn}.
This provides us full control over the quality of assumed neighborhoods, \ie,  the inlier ratios, which is well suited for our analysis.
This also allows us to showcase the robustness of our approach.
Finally, to show the practical feasibility of the attack, we present results with neighborhoods estimated by our learning-based approach.
The point recovery on these estimated neighborhoods reveals private content even though the proposed neighborhood network is only a proof of concept.
As such, it is simple and has limited scalability to a few thousand descriptors whereas point clouds usually involve several hundred thousand descriptors.
Therefore we run this end-to-end evaluation only in 2D. 
This network however shows the potential that a similar architecture could be trained for more descriptors given hardware with enough memory.

\begin{figure*}[t]
  \centering
    \includegraphics[width=0.80\textwidth]{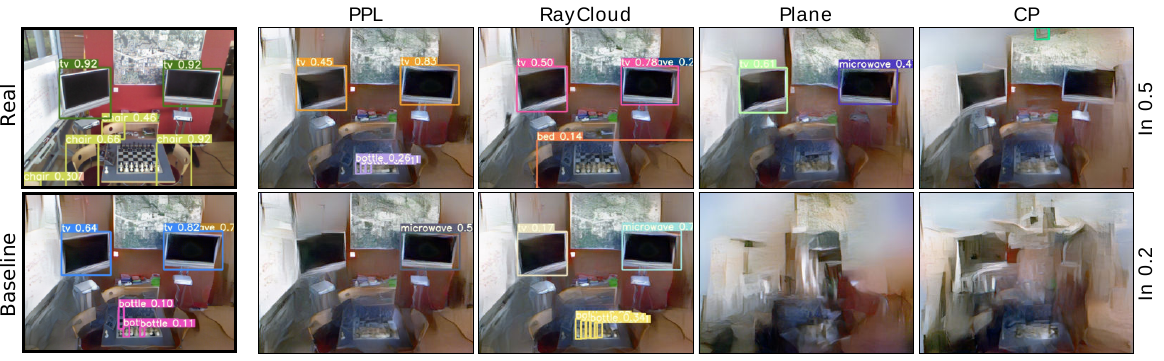}
  \caption{\textbf{Visual content revealed} by the inversion~\cite{pittaluga2019revealing} from the original points (`Baseline') and the points recovered from the 3D obfuscations with neighborhood information at various levels of inlier ratios (In.).
  The original points are triangulated from SIFT~\cite{Lowe04IJCV} features.
  Line obfuscations (OLC)~\cite{speciale2019privacy,speciale2019privacy2d}, Point-Pair-Lines PPL~\cite{lee2023paired} and RayClouds~\cite{moon2024efficient} are more vulnerable to neighborhood-based attacks than Planes~\cite{geppert2022privacy} and Coordinate Permutation~\cite{pan2023privacy}.
  }
  \label{fig:qualitative_3d}
\end{figure*}

\begin{figure*}[t]
  \centering
    \includegraphics[width=0.97\textwidth]{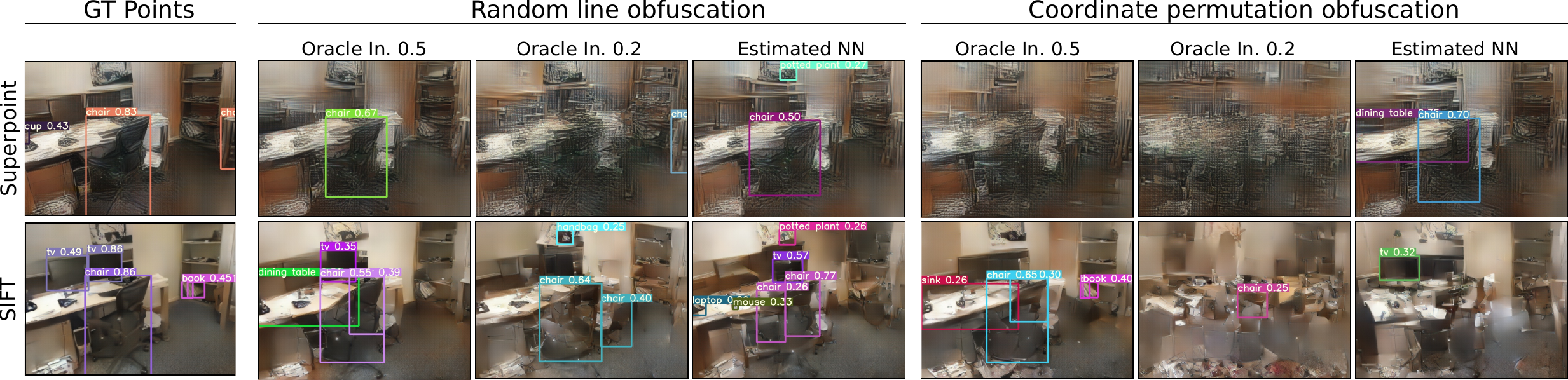}
  \caption{(Best viewed when zoomed in.) \textbf{Visual content revealed} by the inversion applied on points recovered from the obfuscated representations when using two different kinds of keypoints extractors and descriptors - SuperPoint~\cite{detone2018superpoint} and SIFT~\cite{Lowe04IJCV}.
  The columns titled \textit{Estimated NN} show the content revealed with an end-to-end attack, \ie, starting from only descriptors, we carry out neighborhood estimation, point recovery, and inversion to the image space.
  The presence of identifiable scene content in the inverted images emphasizes the vulnerability of current geometry obfuscation techniques.}
  \label{fig:qualitative_2d}
\end{figure*}

\begin{table*}
\centering
\fontsize{8}{6.2}\selectfont
\resizebox{0.85\textwidth}{!}{% Adjust the 0.85 value to control the overall width
\begin{minipage}{0.42\textwidth}
\centering
\begin{tabular}{l*{4}{c}}
\toprule
 & \multicolumn{2}{c}{\textbf{7Scenes.} GT: 0.74} & \multicolumn{2}{c}{\textbf{Cambridge.} GT: 0.53}\\ 
\cmidrule(r){2-3} \cmidrule(r){4-5} 
In. & Lines & CP & Lines  & CP\\ 
\cmidrule(r){2-3} \cmidrule(r){4-5} 
1.0 & 0.62 & 0.62 & 0.40 & 0.41 \\ 
0.5 & 0.62 & 0.58 & 0.40 & 0.37\\ 
0.2 & 0.57 & 0.53 & 0.31 & 0.23\\ 
\bottomrule
\end{tabular}
\label{tab:SSIM_2D}
\end{minipage}%
\hfill
\begin{minipage}{0.56\textwidth}
\centering
\begin{tabular}{l*{8}{c}}
\toprule
 & \multicolumn{4}{c}{\textbf{7Scenes.} GT: 0.58} & \multicolumn{4}{c}{\textbf{Cambridge.} GT: 0.39}\\ 
\cmidrule(r){2-5} \cmidrule(r){6-9} 
In.  & PPL & Plane & CP  & Ray & PPL & Plane & CP & Ray \\ 
\cmidrule(r){2-5} \cmidrule(r){6-9} 
1.0 & 0.57 & 0.55 & 0.56 & 0.57 & 0.36 & 0.36 & 0.36 & 0.37 \\ 
0.5 & 0.56 & 0.49 & 0.51 & 0.56 & 0.36 & 0.32 & 0.34 & 0.37 \\ 
0.2 & 0.54 & 0.43 & 0.43 & 0.54 & 0.36 & 0.31 & 0.27 & 0.37 \\ 
\bottomrule
\end{tabular}
\label{tab:SSIM_3D}
\end{minipage}
}
\caption{\textbf{Perceptual Evaluation} of point recoveries from geometric obfuscations in 2D (left) and 3D (right) with oracle neighborhoods.
The original points are derived from SIFT~\cite{Lowe04IJCV} features.
The SSIM$\uparrow$ compares the original image to the images inverted~\cite{pittaluga2019revealing} from recovered points. {GT} refers to the SSIM of the image inverted from the original points and sets the baseline.
The SSIM for recovered points is in general close to the baseline, demonstrating that the image content is recovered.}
\label{tab:SSIM_results}
\end{table*}

\begin{table*}
\centering
\fontsize{8}{6.2}\selectfont
\resizebox{0.75\textwidth}{!}{% Adjust \textwidth to control overall size
\begin{tabular}{l*{12}{c}}
\toprule
 & \multicolumn{6}{c}{\textbf{Superpoint}~\cite{detone2018superpoint}} & \multicolumn{6}{c}{\textbf{SIFT}~\cite{Lowe04IJCV}}\\ 
\cmidrule(r){2-7} \cmidrule{8-13} 
 & \multicolumn{4}{c}{\textbf{Geometric}} & \multicolumn{2}{c}{\textbf{Perceptual}} & \multicolumn{4}{c}{\textbf{Geometric}} & \multicolumn{2}{c}{\textbf{Perceptual}} \\ 
\cmidrule(r){2-5} \cmidrule(r){6-7} \cmidrule(r){8-11} \cmidrule(r){12-13}  
& \multicolumn{2}{c}{\textbf{Lines}} & \multicolumn{2}{c}{\textbf{CP}} & \textbf{Lines} & \textbf{CP} & \multicolumn{2}{c}{\textbf{Lines}} & \multicolumn{2}{c}{\textbf{CP}} & \textbf{Lines} & \textbf{CP} \\
\cmidrule(r){2-3} \cmidrule(r){4-5} \cmidrule(r){6-6} \cmidrule(r){7-7} \cmidrule(r){8-9} \cmidrule(r){10-11} \cmidrule(r){12-12} \cmidrule(r){13-13}  
 Neighborhood & 10px & 25px & 10px & 25px & \multicolumn{2}{c}{SSIM - GT:0.57} & 10px & 25px & 10px & 25px & \multicolumn{2}{c}{SSIM - GT:0.74} \\
\cmidrule(r){2-3} \cmidrule(r){4-5} \cmidrule(r){6-6} \cmidrule(r){7-7} \cmidrule(r){8-9} \cmidrule(r){10-11} \cmidrule(r){12-12} \cmidrule(r){13-13}
Oracle In. 0.75 & 61.4 & 91.2 & 54.4 & 83.4 & 0.46 & 0.45 & 77.6 & 96.1 & 72.7 & 91.8 & 0.62 & 0.58\\
Oracle In. 0.5 &  63.0 & 92.9 & 45.6 & 70.3 & 0.46 & 0.42 & 78.8 & 96.9 &  63.9 & 80.2 & 0.62 & 0.58\\ 
Oracle In. 0.3 &  57.8 & 87.8 & 23.7 & 38.4 & 0.45 & 0.40 & 73.1 & 92.0 & 32.3 & 43.6 & 0.61 & 0.53\\
Oracle In. 0.2 &  45.1 & 70.8 & 13.3	& 23.1 & 0.40 & 0.40 & 56.9 & 74.8 &  15.9 & 24.6 & 0.57 & 0.53\\
Oracle In. 0.1 &  22.9 & 39.0 & 6.7 & 13.3 & 0.33 & 0.40 & 26.2 & 40.0 &  6.9 & 13.2 & 0.51 & 0.52\\ 
Estimated (Ours) & 53.2 & 86.2 & 48.3 & 80.3 & 0.46 & 0.45 & 47.3 & 68.1 & 30.1 & 45.8 & 0.57 & 0.55 \\
\bottomrule
\end{tabular}%
}
\caption{\textbf{End-to-end attack evaluation}. Geometric and perceptual evaluation of the recovery when using neighbors estimated by our network described in Sec.~\ref{sec:rec_nn} (last row) for two different types of keypoint detectors and extractors—Superpoint~\cite{detone2018superpoint} and SIFT~\cite{Lowe04IJCV}. The performance when using oracle-provided neighborhoods of different qualities is provided for comparison.
The recovery of neighborhoods is observed to be much more effective using Superpoint~\cite{detone2018superpoint} descriptors compared to SIFT~\cite{Lowe04IJCV}.}
\label{tab:results_nn_est}
\end{table*}

\PAR{Implementation details.}
The minimization problem~\eqref{eq:estimate} is formulated as a least-square problem solved using the Ceres solver~\cite{Agarwal_Ceres_Solver_2022} in a RANSAC~\cite{fischler1981random}-like loop. 
Given a set of obfuscated representations, each point is recovered individually using only its neighborhood.
Although this approach does not model the dependencies between the recovery of each point, it allows for a simple parallelization and efficient runtimes, even on a single CPU (see the supp. mat.). %

The oracle based neighborhoods are generated using the original points: a neighborhood of size $K$ with inlier ratio $In.$ is made by first selecting the $K$ nearest neighbor of the original point and replacing $(1-In.)\cdot K$ of them with points randomly chosen from the set of non-neighbors.
The learned neighborhoods are derived as the top-$K$ elements of each row in the similarity matrix output by the network.
The network is trained on the top $K$=20 neighbors of 309K images from 184 Scannet~\cite{dai2017scannet} scenes (see supp. mat.).

% Once the approximate positions of the obfuscated points are recovered, they are fed together with their descriptors to an inversion network~\cite{dusmanu2021privacy,pittaluga2019revealing} to generate images of the scene.
The recovered positions of the obfuscated points, together with the descriptors are fed to an inversion network~\cite{dusmanu2021privacy,pittaluga2019revealing} to generate images of the scene.
% This assumes that the geometric obfuscation preserves the descriptor, which is a valid assumption in the context of visual localization. 
As a valid assumption in the context of visual localization, the descriptor is assumed to be not modified.
This is true for all the obfuscations discussed in this paper except for PPL/PPL+~\cite{lee2023paired}. 
These obfuscations map a pair of points and the corresponding descriptors to the same line, without preserving the mapping between the points and their descriptors.
We again use the neighborhood information to recover this point-descriptor mapping (see supp. mat. for details).

\PAR{Datasets and metrics.}
We evaluate on the two indoor datasets 7-scenes~\cite{shotton2013scene} and 12-scenes~\cite{valentin2016learning}, and the outdoor dataset Cambridge~\cite{kendall2015posenet}.
Results on 12-scenes %
are included in the supp. mat as they follow the same trend as results on 7-scenes. %
Similarly, results for SuperPoint~\cite{detone2018superpoint} are left for supp. mat as they follow a similar trend as the results using SIFT~\cite{Lowe04IJCV}.
We report the geometric accuracy as the fraction of points recovered within chosen error thresholds.
The threshold is in pixels for 2D obfuscations and in cm for 3D. 
Larger thresholds are used for larger (outdoor) scenes in 3D. 
% of 5, 10, and 25 pixels in 2D, and 10, 25, and 50 cm in 3D depending on whether the scene is indoor (small-scale) or outdoor (large-scale). 
We compare the quality of the images generated from the recovered points against the ones generated from the original points by comparing their respective similarities to the real image.
The similarity is computed with standard perceptual metrics: SSIM, PSNR and  LPIPS~\cite{zhang2018perceptual}.
We report the last two metrics in the supp. material.

\PAR{Geometric evaluation.}
The geometric accuracies for 2D and 3D are reported in Tables~\ref{tab:geom_acc_2d}
and~\ref{tab:geom_acc_3d},  respectively.
The proposed generic recovery method can consistently recover the points within a few pixels in the 2D case and within a few cms in 3D maps.
Note that for 2D and 3D line-based obfuscations the performance can peak when the neighborhood is not perfect, \ie, when the inlier ratio is lower than 1.0, although the variation is small.
This is because our robust method identifies outliers in the neighborhood and filters them out and using fewer close points can result in better accuracy using our method. 
% Although outlier-free, such a neighborhood has a different size and different elements compared to the perfect neighborhood with inlier ratio 1.0 so the optimization can estimate slightly different recovered points.

\PAR{Perceptual evaluation.}
All perceptual metrics show consistent results so we report only the SSIM in~\cref{tab:SSIM_3D}.
`GT' is the baseline SSIM between the real image and the one generated from the \textit{original} points, which can be interpreted as an upper bound for the SSIM between the real image and the one generated from the \textit{recovered} points.
The difference from this bound is higher for 2D than for 3D.
This is because even a high geometric error in 3D can reduce to very few pixels upon projection while the recovery from obfuscations in 2D leads to several pixels of error.
The larger error in 2D keypoint position estimation leads to worse image reconstructions in case of 2D obfuscations.
% This larger error in the sparse set of keypoint locations, which is an input to the inversion network, leads to a larger gap in perceptual metrics.

\PAR{Learned neighborhoods.}
We evaluate the 2D point recovery from lines~\cite{speciale2019privacy2d} and CP~\cite{pan2023privacy} using learned neighborhoods. %
We use the top-K=20 neighbors derived from the similarity output by our network from Sec.\ref{sec:rec_nn}.
~\cref{tab:results_nn_est} shows the geometric and perceptual performance of an end-to-end attack, while \cref{fig:qualitative_2d} shows the inverted images.

An interesting observation is that the network learns the neighborhood more easily for SuperPoint~\cite{detone2018superpoint} than for SIFT~\cite{Lowe04IJCV} as indicated by the accuracy gap between the two: for SuperPoint~\cite{detone2018superpoint}, the network leads to neighborhoods with acc. between 70\% and 80\% for $K \in [10, 100] $ while for SIFT~\cite{Lowe04IJCV}, the accuracy remains around 35\%.
Thus one could argue that SIFT is more privacy-preserving than SuperPoint, although there is no guarantee that better neighborhood estimators for SIFT will not become available in the future.
Moreover, SIFT~\cite{Lowe04IJCV} typically achieves lower localization performance than SuperPoint~\cite{sattler2018benchmarking} and sacrificing performance for privacy might not be a satisfying solution in all scenarios.

Even though the images inverted from the recovered points are not perfect, the outline and the objects in the scene are recognizable.
These results highlight an important limitation of pure geometric obfuscations and support the two claims made in the paper:
i) the neighborhood information can be learned from the descriptors, and reiterates that geometric obfuscations alone are not as privacy-preserving as they claim.
One needs to also prevent neighborhood information from being inferred from the obfuscations.
ii) it shows that the proposed proof of concept to compute the neighborhood information is already sufficient for the proposed point recovery to be applicable.
We expect that more complex neighborhood learning will lead to better results.
% This calls for further work on obfuscating the descriptors to prevent learning neighborhoods or other private content, and fusing geometric and descriptors obfuscations.
This calls for potential future work on fusing geometric and descriptor obfuscation to prevent neighborhood recovery.

\PAR{Discussion.}
% The results reveal that the proposed recovery method performs well even if the neighborhoods contain significant fractions of outliers.
% \cref{fig:qualitative_3d} and \cref{fig:qualitative_2d} further show that images generated from recovered points can reveal potentially private user content, which is particularly true for line-based obfuscations~\cite{speciale2019privacy,speciale2019privacy2d,lee2023paired}.
% The geometric constraints of parallel planes~\cite{geppert2022privacy} make recovery difficult, but neighborhoods with reasonable inlier ratios make the plane obfuscation also susceptible to the proposed recovery.
% The same holds for Coordinate Permutation~\cite{pan2023privacy}: 
% neighborhoods with inlier ratios of 0.5 or more are enough to enable recovery accurate enough to reveal identifiable scene content.
% The network described in~\cref{sec:rec_nn} can produce such informative neighborhoods even with its simple design.
% We expect methods in future works to improve the estimation of neighborhoods from descriptors, further highlighting the discussed vulnerability of obfuscation schemes.
The results reveal that the proposed recovery method performs well even if the neighborhoods contain significant fractions of outliers.
\cref{fig:qualitative_3d} and \cref{fig:qualitative_2d} further show that images generated from recovered points can reveal potentially private user content, which is particularly true for line-based obfuscations~\cite{speciale2019privacy,speciale2019privacy2d,lee2023paired}.
The geometric constraints of parallel planes~\cite{geppert2022privacy} make recovery difficult, but neighborhoods with reasonable inlier ratios make the plane obfuscation also susceptible to the proposed recovery.
The same holds for Coordinate Permutation~\cite{pan2023privacy}: the additional step of estimating which coordinate was permuted brings in more noise into our method for recovering points.
However, neighborhoods with inlier ratios of 0.5 or more are enough to enable recovery accurate enough to reveal identifiable scene content.
The network described in~\cref{sec:rec_nn} can produce such informative neighborhoods even with its simple design.
We expect methods in future works to improve the estimation of neighborhoods from descriptors, further highlighting the discussed vulnerability of obfuscation schemes.
Future methods in de-noising the neighborhood graphs estimated from descriptor and/or geometry can help reduce the error in point position recovery.
Similarly, more sophisticated inversion attacks that are robust to small noise in point positions can increase the privacy risk.

\section{Conclusion}
In this work, we highlight a common vulnerability of all geometry-based obfuscation techniques that have so far been presented as privacy-preserving representations.
We present a simple optimization-based method that uses knowledge of point neighborhoods to recover point positions from the discussed obfuscation schemes.
We show the robustness of our method and analyze the recovery accuracy by using oracle-provided neighborhoods with varying inlier ratios.
Finally, using a neural network that learns to identify local feature descriptors co-occurring across scenes, we show that it is possible to estimate these neighborhoods from the descriptors associated with points.
The inverted images from the recovered point positions reveal private scene content, highlighting the drawback of current methods and the need for %
guarantees on under which circumstances a data representation is indeed privacy-preserving. %

\PAR{Acknowledgements.}
This work was supported by 
% Zuzana
the Czech Science Foundation (GACR) JUNIOR STAR Grant No. 22-23183M, 
% BEGINNING OF EDIT
% PREVIOUSLY (BEFORE MONDAY 10th February)
%% Assia and Torsten
%the EU Horizon 2020 project RICAIP (No. 857306),
% AFTER
% Torsten
the EU Horizon 2020 project RICAIP (No. 857306), 
% END OF EDIT
% Kunal and Fredrik
the Chalmers AI Research Center (CHAIR), WASP and SSF.
The compute and storage were partially supported by NAISS projects numbered NAISS-2024/22-637 and NAISS-2024/22-237 respectively.

% {
%     \small
%     \bibliographystyle{ieeenat_fullname}
%     \bibliography{main}
% }

% \clearpage
\appendix
% \title{Obfuscation Based Privacy Preserving Representations are Recoverable Using Neighborhood Information - Supplementary Material}

\maketitlesupplementary

The supplementary material is organized as follows.
Sec.~\ref{supp:swap_coord_est} details the point recovery from coordinate permutations~\cite{pan2023privacy} and how we estimate which of the coordinates is swapped to transform the recovery from coordinate permutations into a recovery from lines.
Sec.~\ref{supp:desc_assignment} recalls the descriptor ambiguity in paired-point lines obfuscations~\cite{lee2023paired} and how neighborhood information is used to assign descriptors to their original points.
Sec.~\ref{supp:additional_results} reports results on the indoor 12-scenes~\cite{valentin2016learning} dataset as announced in Section 6.
These results are consistent with the ones on the indoor 7-scenes~\cite{shotton2013scene} dataset.
We also report the geometric and perceptual evaluation for all 3D obfuscations, including the random line obfuscation OLC~\cite{speciale2019privacy} and the PPL+ variant of the pair-point lifting~\cite{lee2023paired}, that are left out of the main paper for the sake of brevity.
Additionally, we also provide visual examples of the estimated neighborhood graph on two scenes from the ScanNet++~\cite{yeshwanthliu2023scannetpp} dataset.
Sec.~\ref{supp:impl_details} provides additional implementation details related to the nearest-neighbor learning and the image inversion from 2D points.

\section{Coordinate permutation - Predicting swapped coordinate}
\label{supp:swap_coord_est}

As mentioned in Sec.4 of the paper, the coordinate permutation obfuscation is equivalent to obfuscating the points with multiple lines (2 in 2D, 3 in 3D) that are axes-aligned and pass through the obfuscated point $\mathcal{O}(x)$.
It should be recalled that this is done for the computational feasibility of the proposed approach as explained in the main paper.
Before running the proposed recovery method on these lines, we discard some of the lines so that for each point, only one of the two or three lines remains.
The remaining line should follow the direction along which the point has been moved.
Identifying such a line amounts to estimating which of the coordinates of the obfuscated points have been swapped.
We now describe how to identify such a line, \ie, how to identify the swapped coordinate.

\begin{figure}
    \centering
    \includegraphics[width=0.9  \linewidth]{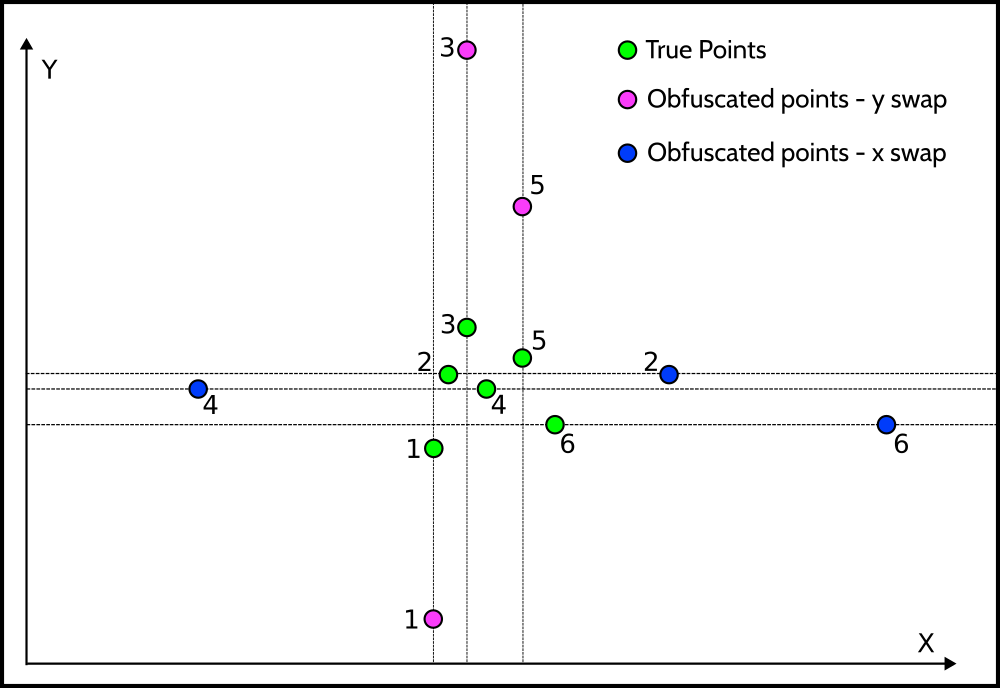}
    \caption{\small{\textbf{Illustration of the Coordinate Swap Inversion.} The green points represent the true original points that form a neighborhood. One coordinate of each point is swapped with that of another point in the image (not shown here for brevity) to result in the blue/pink points. 
    Note that points shifted along the y-axis (pink) form a cluster around the same x-value and similarly points shifted along the x-axis form a cluster around the same y-value. This idea is used to estimate the swapped coordinates of the members of a neighborhood.}}
    \label{fig:swap_coord_class_course}
\end{figure}

For each point, we predict the swapped coordinate, correspondingly the line along which the original point is estimated to lie, using neighborhood information.
Our method, as illustrated in Fig.~\ref{fig:swap_coord_class_course}, is based on the observation that if one arbitrary coordinate of the points in a neighborhood is changed, then the obfuscated points (ones with swapped coordinates) remain close to each other along the remaining dimensions (the coordinates that were not swapped).
In practice, given a set of obfuscated points that are known to be neighbors, we iterate through each point and compute its distances to all other points in the neighborhood along each axis.
We identify the axis which has relatively larger cumulative distances as the estimated line direction.
For example, in Fig.~\ref{fig:swap_coord_class_course}, the green points show the original points in a neighborhood, and the set of blue and pink points together form the set $\mathcal{N}(\mathcal{O}(x))$ of obfuscated neighbors. 
Then, if we consider the blue point numbered 4, it has small distances along the y-axis to the blue points 2 and 6 but relatively larger distances along the x-axis to all points. 
It is therefore estimated to have been moved along the x-axis.
We make this approach robust with voting, \ie, we visit the neighborhood of all points and accumulate the estimated direction for each point over all visits. 
In the end, we select the direction with the most votes for each point.

\section{Descriptor Assignment for Paired-Point-Lines}
\label{supp:desc_assignment}

The point-paired 3D line obfuscations, PPL and PPL+~\cite{lee2023paired}, transform the point cloud into a line cloud by generating lines joining random pairs of 3D points.
This approach has several advantages one of which is the confusion over feature descriptors.
With the line joining two points, it also holds two descriptors, each associated with one point. 
While the neighborhood-based recovery estimates the position of the two original points on the line, the descriptors still need to be assigned to each of the points to enable the inversion attack~\cite{pittaluga2019revealing}.
We provide a more formal definition of the problem and its solution.

\textbf{Problem Definition:} 
In the paired-point setting, one line holds two descriptors and the point recovery relies on two sets of neighboring lines (one for each obfuscated point).
Each set of neighboring lines is used independently to estimate the position of one obfuscated point.
We then want to associate each estimated point with one of the two descriptors on the line.

\textbf{Solution:} We first note that there is a bijection between an estimated point and a set of neighboring lines.
Assigning a descriptor to an estimated point is then equivalent to assigning a descriptor to a \textbf{set} of neighboring lines.
The intuition behind the proposed method is to assign each descriptor to one of the two sets of neighboring lines.
To choose between the possible assignments, we assign the descriptor to the most `similar' set of neighboring lines, \ie, the set of lines with the most similar descriptors.
We define the distance between a descriptor and a set of neighboring lines as the sum of the distances between the descriptor to be assigned and the descriptor of each neighboring line.
To deal with the fact that the neighboring lines also hold two descriptors, we chose to only count the distance to the closest of the two descriptors of a given neighboring line.
In practice, we compute 4 such distances between each of the two descriptors to be assigned and each of the two sets of neighboring lines.
Each descriptor is assigned to one set so that the cumulative distance of the assignment is minimized.

Note that the derivation only takes as input the lines, the pair of descriptors on each line, and the neighborhood set. The position of the points, whether original or estimated, is never used.

\section{Additional Results}
\label{supp:additional_results}

As a reminder, the geometric evaluation measures how close the points recovered from the obfuscations are to the original points.
We measure the accuracy of the recovered points as the ratio of points which Euclidean distance to the original ones is below a given threshold in cm in 3D, and in pixels in 2D.
The perceptual evaluation measures how close the images inverted~\cite{pittaluga2019revealing} from the points recovered from the obfuscations are to the images inverted from the original points.
We report three metrics that measure the similarity between images: the Structural Similarity Index Measure (SSIM), the Peak-to-Signal Noise Ratio (PSNR), and the Learned Perceptual Image Patch Similarity~\cite{zhang2018perceptual} (LPIPS).

\PAR{Geometric Evaluation of obfuscations in 2D.}
Tables \ref{supp:tab_geom_12scenes_2d} and \ref{supp:tab_perceptual_12scenes_2d} show results over the 12-scenes~\cite{valentin2016learning} dataset using SuperPoint~\cite{detone2018superpoint} and SIFT~\cite{Lowe04IJCV} as the local features.
The results using SIFT on 12-scenes~\cite{valentin2016learning} follow the same trend as the results for 7-scenes~\cite{shotton2013scene} shown in the main paper, with a relative difference of 1-8\% in geometric accuracy.
However, the geometric accuracy of the recovered points is slightly lower when using Superpoint~\cite{detone2018superpoint} on 12-scenes~\cite{valentin2016learning}.
We believe that this is because the keypoints are more sparsely distributed on these images: i) the 12-scenes images are larger (1296x968) than the 7-scenes~\cite{shotton2013scene} ones (640x480); ii)
SuperPoint~\cite{detone2018superpoint} features are typically much sparser than the SIFT~\cite{Lowe04IJCV} features, leading to a larger distance between an obfuscated point and the points in the neighborhood.
Since the distance between an obfuscated points and its furthest neighbor is an upper bound on the error of the recovered point~\cite{chelani2021privacy}, a larger mean distance to the neighbors usually implies a decrease in the geometric accuracy.
This suggests that one way to prevent the proposed point recovery is to use sparse keypoints but this may come at the cost of lower localization performance.
Also, we observe that using a smaller neighborhood size improves the accuracy for SuperPoint~\cite{detone2018superpoint} so sparsifying the points may not be enough since tuning the parameters of the point recovery can compensate for it.
To keep the recovery parameters consistent with the rest of the paper, we show all results for 2D obfuscations using $K=20$ as the neighborhood size.

\begin{figure*}[t]
  \centering
    \includegraphics[width=0.9\linewidth]{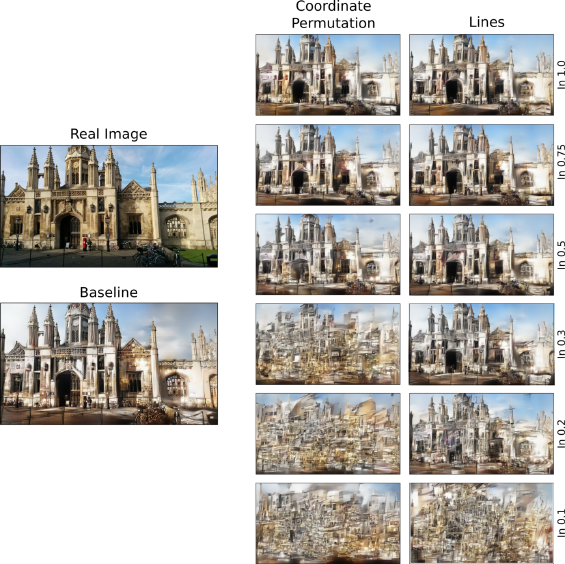}
  \caption{\textbf{Additional Qualitative Results - Cambridge~\cite{shotton2013scene} dataset, scene \textit{King's College}}.
  Images inverted from the original 2D SIFT~\cite{Lowe04IJCV} keypoints and the points recovered from the 2D obfuscations using neighborhood information with various levels of inlier ratios (In.).
  }
\label{supp:quali_2D_kings}
\end{figure*}

\begin{figure*}[t]
  \centering
    \includegraphics[width=0.9\linewidth]{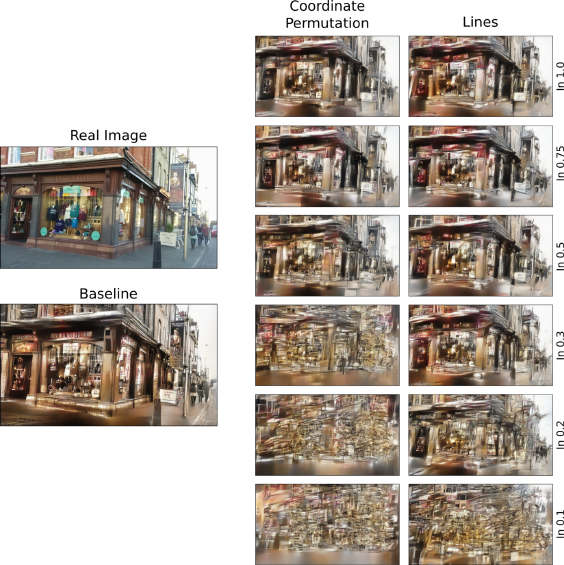}
  \caption{\textbf{Additional Qualitative Results - Cambridge~\cite{kendall2015posenet} dataset, scene \textit{Shop Facade}}.
  Images inverted from the original 2D SIFT~\cite{Lowe04IJCV} keypoints and the points recovered from the 2D obfuscations using neighborhood information with various levels of inlier ratios (In.).
  }
\label{supp:quali_2D_shop_facade}
\end{figure*}

\begin{figure*}[t]
  \centering
    \includegraphics[width=0.9\linewidth]{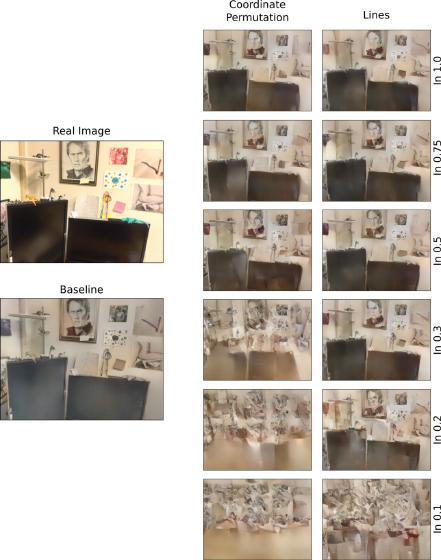}
  \caption{\textbf{Additional Qualitative Results - 12scenes~\cite{valentin2016learning} dataset, scene \textit{Office1-manolis}}.
  Images inverted from the SIFT~\cite{Lowe04IJCV} descriptors and original 2D keypoints and the points recovered from the 2D obfuscations using neighborhood information with various levels of inlier ratios (In.).
  }
\label{supp:quali_2D_12sc_office1_manolis}
\end{figure*}

\begin{figure*}[t]
  \centering
    \includegraphics[width=0.9\linewidth]{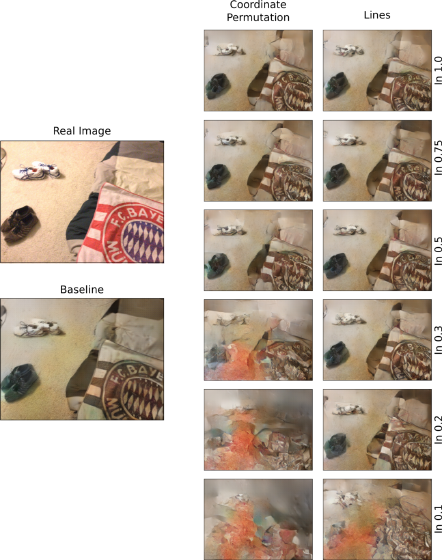}
  \caption{\textbf{Additional Qualitative Results - 12scenes~\cite{valentin2016learning} dataset, scene \textit{Apt2-bed}}.
  Images inverted from the SIFT~\cite{Lowe04IJCV} descriptors and original 2D keypoints and the points recovered from the 2D obfuscations using neighborhood information with various levels of inlier ratios (In.).
  }
\label{supp:quali_2D_12sc_apt2_bed}
\end{figure*}

\PAR{Perceptual Evaluation of obfuscations in 2D.}
Figures \ref{supp:quali_2D_12sc_office1_manolis} and \ref{supp:quali_2D_12sc_apt2_bed} show qualitative results for images inverted in indoor scenes when using oracle-provided neighborhoods of different qualities and SIFT~\cite{Lowe04IJCV} features.
It is clear that identifiable scene content is revealed even for neighborhoods of inlier ratio 0.2 in case of lifting to random lines~\cite{speciale2019privacy2d}. 
With coordinate permutations~\cite{pan2023privacy}, the scene remains more private and we observe that the performance bottleneck of the point recovery lies in the preprocessing step that estimates which coordinate is swapped using the neighborhood information.
Still, neighborhoods with inlier ratios of 0.5 or more are enough for the point recovery to successfully reveal the content of the scene.
Figures \ref{supp:quali_2D_kings} and \ref{supp:quali_2D_shop_facade} show similar results for outdoor scenes from the Cambridge~\cite{kendall2015posenet} dataset.

\begin{table*}[t]
\begin{center}
\begin{tabular}{l*{12}{c}}
\toprule
& \multicolumn{6}{c}{\textbf{SuperPoint}~\cite{detone2018superpoint}}
& \multicolumn{6}{c}{\textbf{SIFT}~\cite{Lowe04IJCV}} \\
\cmidrule(r){2-7} \cmidrule(r){8-13}
& \multicolumn{3}{c}{\textbf{CP}~\cite{pan2023privacy}}
& \multicolumn{3}{c}{\textbf{Lines}~\cite{speciale2019privacy2d}}
& \multicolumn{3}{c}{\textbf{CP}~\cite{pan2023privacy}} 
& \multicolumn{3}{c}{\textbf{Lines}~\cite{speciale2019privacy2d}} \\
\cmidrule(r){2-4} \cmidrule(r){5-7} \cmidrule(r){8-10} \cmidrule(r){11-13} 
In. & 5px & 10px & 25px & 5px & 10px & 25px & 5px & 10px & 25px & 5px & 10px & 25px \\ 
\cmidrule(r){2-4} \cmidrule(r){5-7} \cmidrule(r){8-10} \cmidrule(r){11-13} 
1.0  & 12.9 & 24.43 & 52.8  & 13.5 & 26.4  & 56.6  & 42.7 & 67.2  & 89    & 41.3 & 66.8  & 89.6 \\ 
0.75 & 14.5 & 27.4  & 56.5  & 15.8 & 30.7  & 62.42 & 43.8 & 69.1  & 89.7  & 42.9 & 69.6  & 91.5 \\ 
0.50 & 14.4 & 26.4  & 52.4  & 18.5 & 35.6  & 69.1  & 40   & 64    & 82.1  & 43.7 & 71.7  & 93.3 \\ 
0.30 & 8.68 & 15.98 & 30.72 & 19.7 & 38.0    & 71.6  & 20.1 & 33.1  & 43.9  & 41.5 & 69.8  & 91.9 \\ 
0.20 & 4.93 & 9.1   & 17.7  & 17.5 & 33.8  & 63.0    & 9.0    & 14.9  & 21.8  & 34.2 & 58.6  & 79.0   \\ 
0.10 & 2.35 & 4.39  & 9.06  & 9.09 & 17.4  & 32.7  & 3.2  & 5.5   & 9.7   & 15   & 25.5  & 37.7 \\
\bottomrule
\end{tabular}
\caption{\textbf{Geometric accuracy of the point recovery from 2D obfuscations} on the 12-scenes~\cite{valentin2016learning} dataset using two different features : SuperPoint~\cite{detone2018superpoint} and SIFT~\cite{Lowe04IJCV}.
The geometric accuracy, \ie, the fraction of recovered points with an error lower than a given threshold, is lower in general as compared to results over 7Scenes~\cite{shotton2013scene} because of the larger image sizes in the 12-scenes~\cite{valentin2016learning} dataset - 1296x968 as compared to 640x480.
Further, SuperPoint~\cite{detone2018superpoint} features are typically much sparser than SIFT features, increasing the average distance to neighbors.
The average number of SuperPoint~\cite{detone2018superpoint} features per image in our experiment was around 312 as compared to 1412 for SIFT~\cite{Lowe04IJCV}.
This suggests that one way to prevent the proposed point recovery is to use sparse keypoints but this may come at the cost of a lower localization performance.
}
\label{supp:tab_geom_12scenes_2d}
\end{center}
\end{table*}

\begin{table*}[t]
\begin{center}
\begin{footnotesize} %
\begin{tabular}{l*{12}{c}}
\toprule
& \multicolumn{6}{c}{\textbf{SuperPoint}~\cite{detone2018superpoint}}
& \multicolumn{6}{c}{\textbf{SIFT}~\cite{Lowe04IJCV}} \\
\cmidrule(r){2-7} \cmidrule(r){8-13}
& \multicolumn{3}{c}{\textbf{CP}~\cite{pan2023privacy}}
& \multicolumn{3}{c}{\textbf{Lines}~\cite{speciale2019privacy2d}}
& \multicolumn{3}{c}{\textbf{CP}~\cite{pan2023privacy}}
& \multicolumn{3}{c}{\textbf{Lines}~\cite{speciale2019privacy2d}} \\
\cmidrule(r){2-4} \cmidrule(r){5-7} \cmidrule(r){8-10} \cmidrule(r){11-13} 
In. & \textbf{SSIM$\uparrow$} & \textbf{LPIPS$\downarrow$} & \textbf{PSNR$\uparrow$} &  \textbf{SSIM$\uparrow$} & \textbf{LPIPS$\downarrow$} & \textbf{PSNR$\uparrow$} & \textbf{SSIM$\uparrow$} & \textbf{LPIPS$\downarrow$} & \textbf{PSNR$\uparrow$}  & \textbf{SSIM$\uparrow$} & \textbf{LPIPS$\downarrow$} & \textbf{PSNR$\uparrow$} \\ 
\cmidrule(r){2-4} \cmidrule(r){5-7} \cmidrule(r){8-10} \cmidrule(r){11-13} 
\textbf{Baseline} & 0.55 & 0.48 & 15.5 & 0.55 & 0.48 & 15.5 & 0.60 & 0.55 & 14.6 & 0.60 & 0.55 & 14.6 \\
\cmidrule(r){2-13}
1.0 & 0.42 & 0.6 & 13.5 & 0.42 & 0.59 & 13.8 & 0.52 & 0.63 & 14 & 0.51 & 0.64 & 14.1 \\ 
0.75 &  0.41 & 0.60 & 13.4 & 0.42 & 0.59 & 13.9 & 0.51 & 0.64 & 13.9 & 0.51 & 0.64 & 14.1 \\ 
0.50 & 0.39 & 0.61 & 12.9 & 0.43 & 0.58 & 14.1 & 0.49 & 0.66 & 13.3 & 0.52 & 0.63 & 14.3 \\ 
0.30 & 0.37 & 0.63 & 12.2 & 0.43 & 0.58 & 14.0 & 0.43 & 0.70 & 12.0 & 0.52 & 0.64 & 14.2 \\ 
0.20 & 0.37 & 0.64 & 11.9 & 0.41 & 0.6 & 13.7 & 0.41 & 0.72 & 11.5 & 0.49 & 0.65 & 13.6\\ 
0.10 & 0.36 & 0.65 & 11.6 & 0.35 & 0.63 & 12.8 & 0.41 & 0.73 & 11.2 & 0.41 & 0.70 & 11.9 \\
\bottomrule
\end{tabular}
\end{footnotesize}

\caption{\textbf{Perceptual accuracy of the point recovery from 2D obfuscations} on the 12-scenes~\cite{valentin2016learning} dataset using SuperPoint~\cite{detone2018superpoint} and SIFT~\cite{Lowe04IJCV}.
\textbf{Baseline} refers to the similarity score between the real image and the image inverted from the original points.
The results follow the same trends as that for 7-scenes~\cite{shotton2013scene} shown in the main paper.}
\label{supp:tab_perceptual_12scenes_2d}
\end{center}
\end{table*}

\PAR{Geometric Evaluation of other 3D obfuscations.}
In the main paper, we report results only for a subset of 3D geometric obfuscations because of the page limits: the paired-point lines PPL~\cite{lee2023paired}, the Ray clouds~\cite{moon2024efficient}, the plane obfuscation~\cite{geppert2022privacy} and the point permutation~\cite{pan2023privacy}.
We complete these results with the evaluation of the random-line obfuscation~\cite{speciale2019privacy} and the PPL+ variant of the paired-point lines~\cite{lee2023paired} on the two indoor datasets, 7-scenes~\cite{shotton2013scene} and 12-scenes~\cite{valentin2016learning} and the outdoor dataset Cambridge~\cite{kendall2015posenet} in Tables~\ref{supp:tab_geom_7scenes},~\ref{supp:tab_geom_12scenes},~\ref{supp:tab_geom_cambridge}.
The 3D models are generated with Structure-from-Motion~\cite{schonberger2016structure} from SIFT~\cite{Lowe04IJCV} features, except for 7-scenes~\cite{shotton2013scene} for which additional comparisons are run with the learning-baed SuperPoint~\cite{detone2018superpoint} features.

\begin{figure*}[t]
  \centering
    \includegraphics[width=0.5\linewidth,height=\textheight,keepaspectratio]{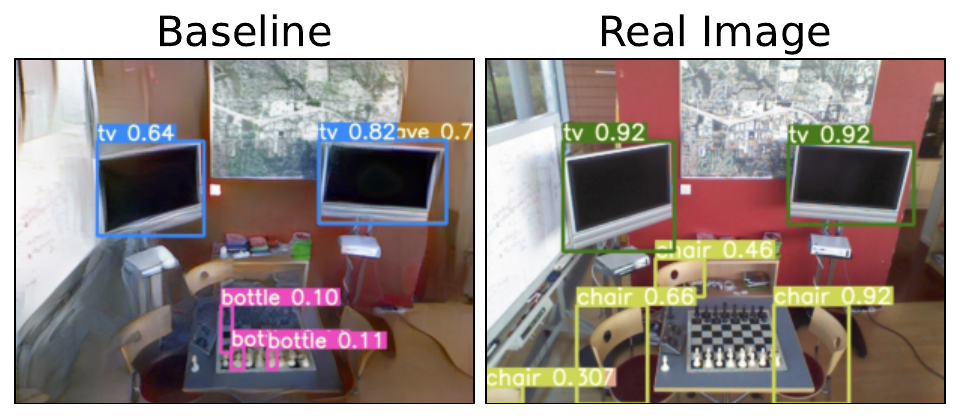}
    \includegraphics[width=\linewidth,height=\textheight,keepaspectratio]
    {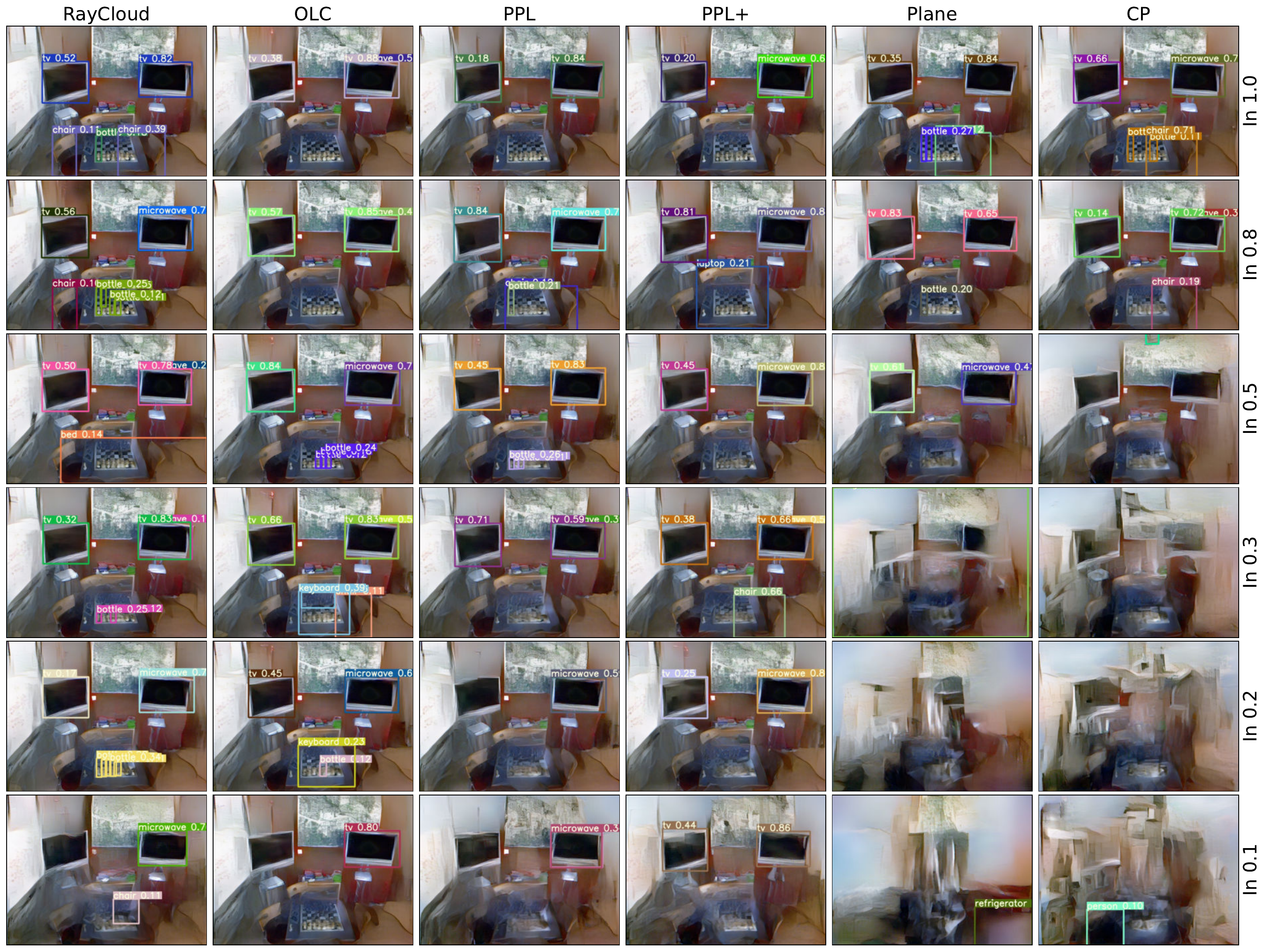}
  \caption{\textbf{Additional Qualitative Results - 7-scenes~\cite{shotton2013scene}-Chess}.
  Images inverted~\cite{pittaluga2019revealing} from the original points (`Baseline') and the points recovered from the 3D obfuscations from neighborhood information with various levels of inlier ratios (In.).
  Line obfuscations (OLC)~\cite{speciale2019privacy,speciale2019privacy2d}, Point-Pair-Lines PPL and PPL+~\cite{lee2023paired}, and ray clouds~\cite{moon2024efficient} are the most vulnerable to neighborhood-based attacks while Planes~\cite{geppert2022privacy} and Permutations~\cite{pan2023privacy} are more privacy preserving. The 3D points cloud is generated from SfM~\cite{schonberger2016structure} on SIFT~\cite{Lowe04IJCV} features.
  }
  \label{supp:quali_3D_chess}
\end{figure*}

\begin{figure*}[t]
  \centering
    \includegraphics[width=0.5\linewidth,height=\textheight,keepaspectratio]{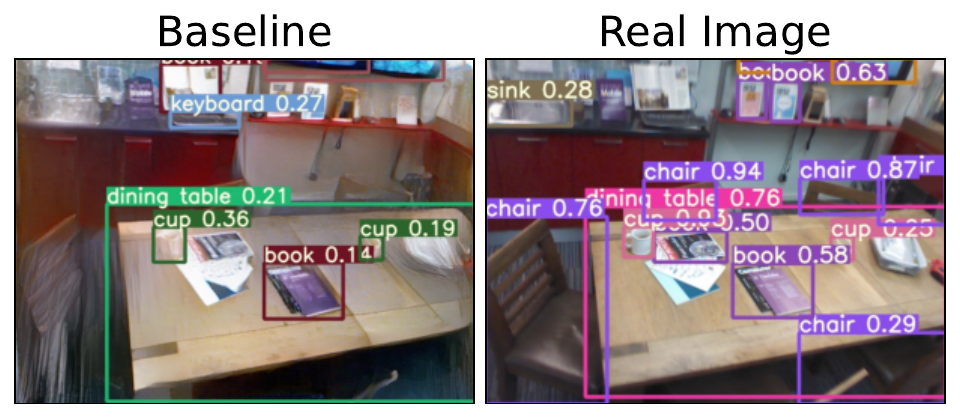}
    \includegraphics[width=\linewidth,height=\textheight,keepaspectratio]
    {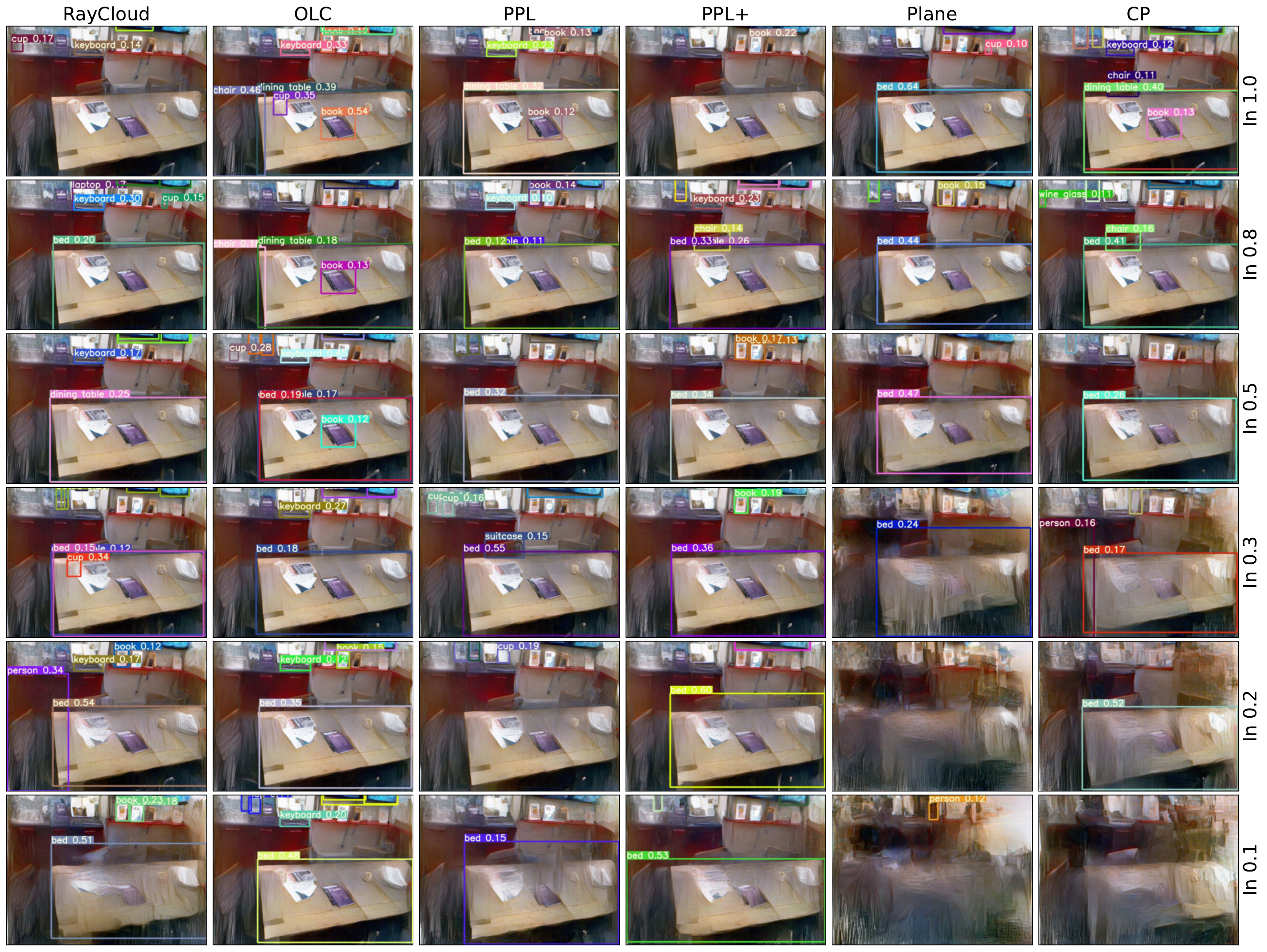}
  \caption{\textbf{Additional Qualitative Results - 7-scenes~\cite{shotton2013scene}-Redkitchen}.
  Images inverted~\cite{pittaluga2019revealing} from the original points (`Baseline') and the points recovered from the 3D obfuscations from neighborhood information with various levels of inlier ratios (In.).
  Line obfuscations (OLC)~\cite{speciale2019privacy,speciale2019privacy2d}, Point-Pair-Lines PPL and PPL+~\cite{lee2023paired} and ray clouds~\cite{moon2024efficient} are the most vulnerable to neighborhood-based attacks while Planes~\cite{geppert2022privacy} and Permutations~\cite{pan2023privacy} are more privacy preserving. The 3D points cloud is generated from SfM~\cite{schonberger2016structure} on SIFT~\cite{Lowe04IJCV} features.
  }
  \label{supp:quali_3D_redkitchen}
\end{figure*}

\begin{figure*}
  \centering
    \includegraphics[width=0.5\linewidth,height=\textheight,keepaspectratio]
    {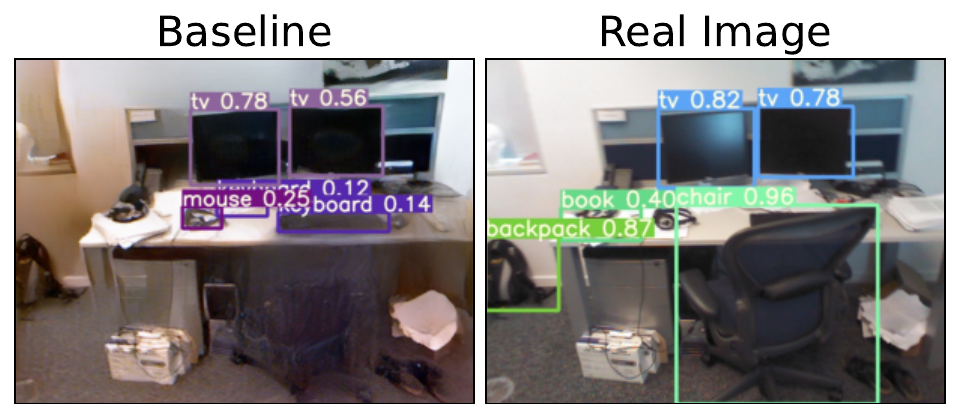}
    \includegraphics[width=\linewidth,height=\textheight,keepaspectratio]
    {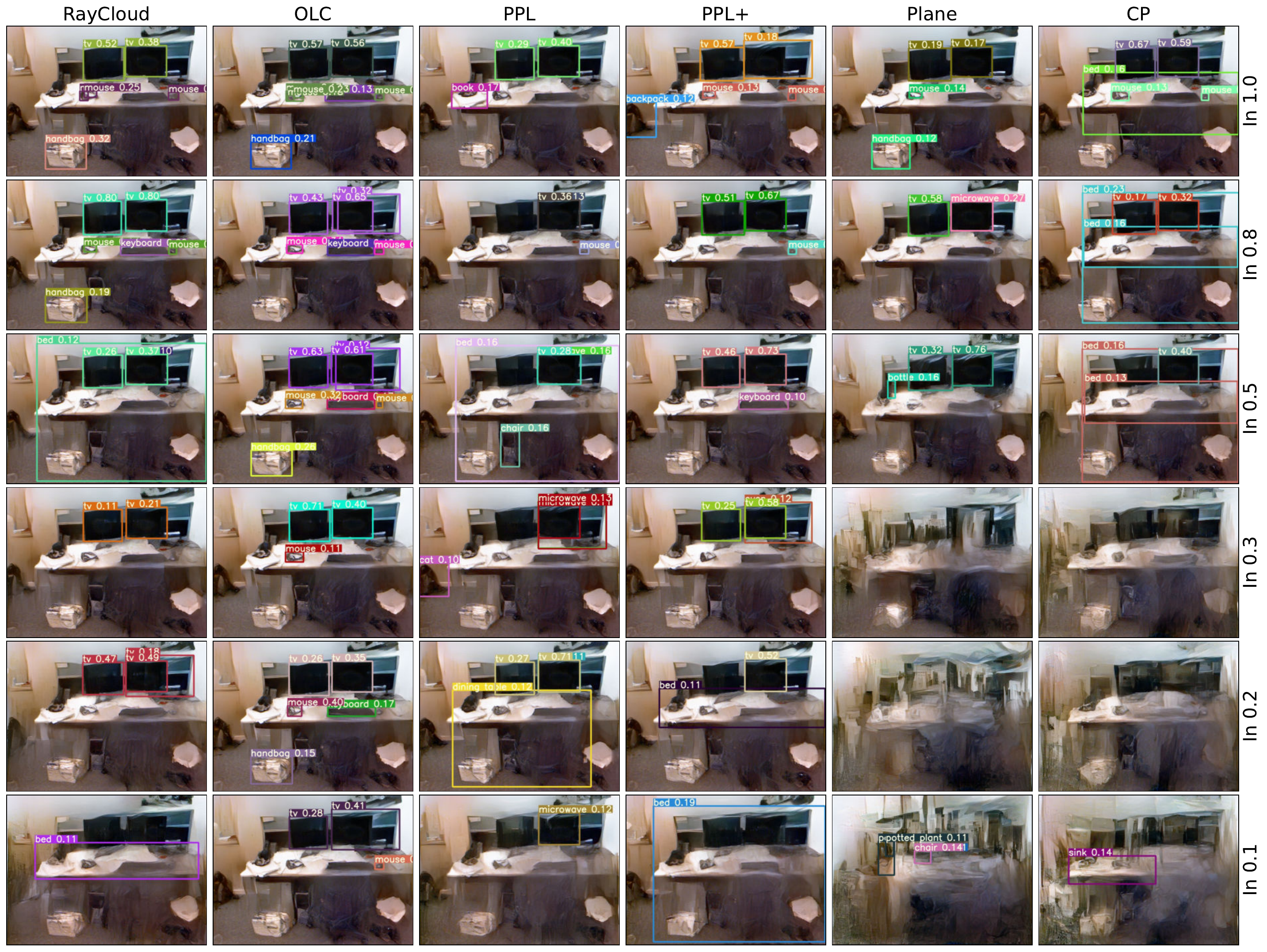}
  \caption{\textbf{Additional Qualitative Results - 7-scenes~\cite{shotton2013scene}-Office}.
  Images inverted~\cite{pittaluga2019revealing} from the original points (`Baseline') and the points recovered from the 3D obfuscations from neighborhood information with various levels of inlier ratios (In.).
  Line obfuscations (OLC)~\cite{speciale2019privacy,speciale2019privacy2d}, Point-Pair-Lines PPL and PPL+~\cite{lee2023paired}, and ray clouds~\cite{moon2024efficient} are the most vulnerable to neighborhood-based attacks while Planes~\cite{geppert2022privacy} and Permutations~\cite{pan2023privacy} are more privacy preserving. The 3D points cloud is generated from SfM~\cite{schonberger2016structure} on SIFT~\cite{Lowe04IJCV} features.
  }
    \label{supp:quali_3D_office}
\end{figure*}

\begin{figure*}
  \centering
    \includegraphics[width=0.5\linewidth,height=\textheight,keepaspectratio]{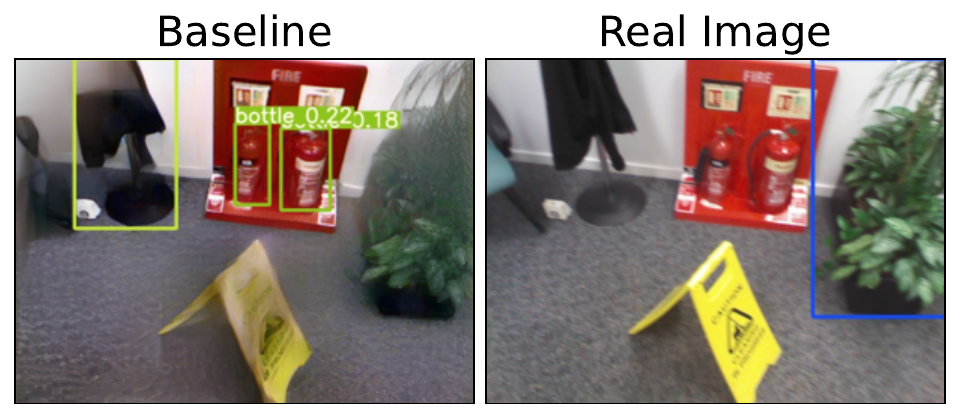}
    \includegraphics[width=\linewidth,height=\textheight,keepaspectratio]
    {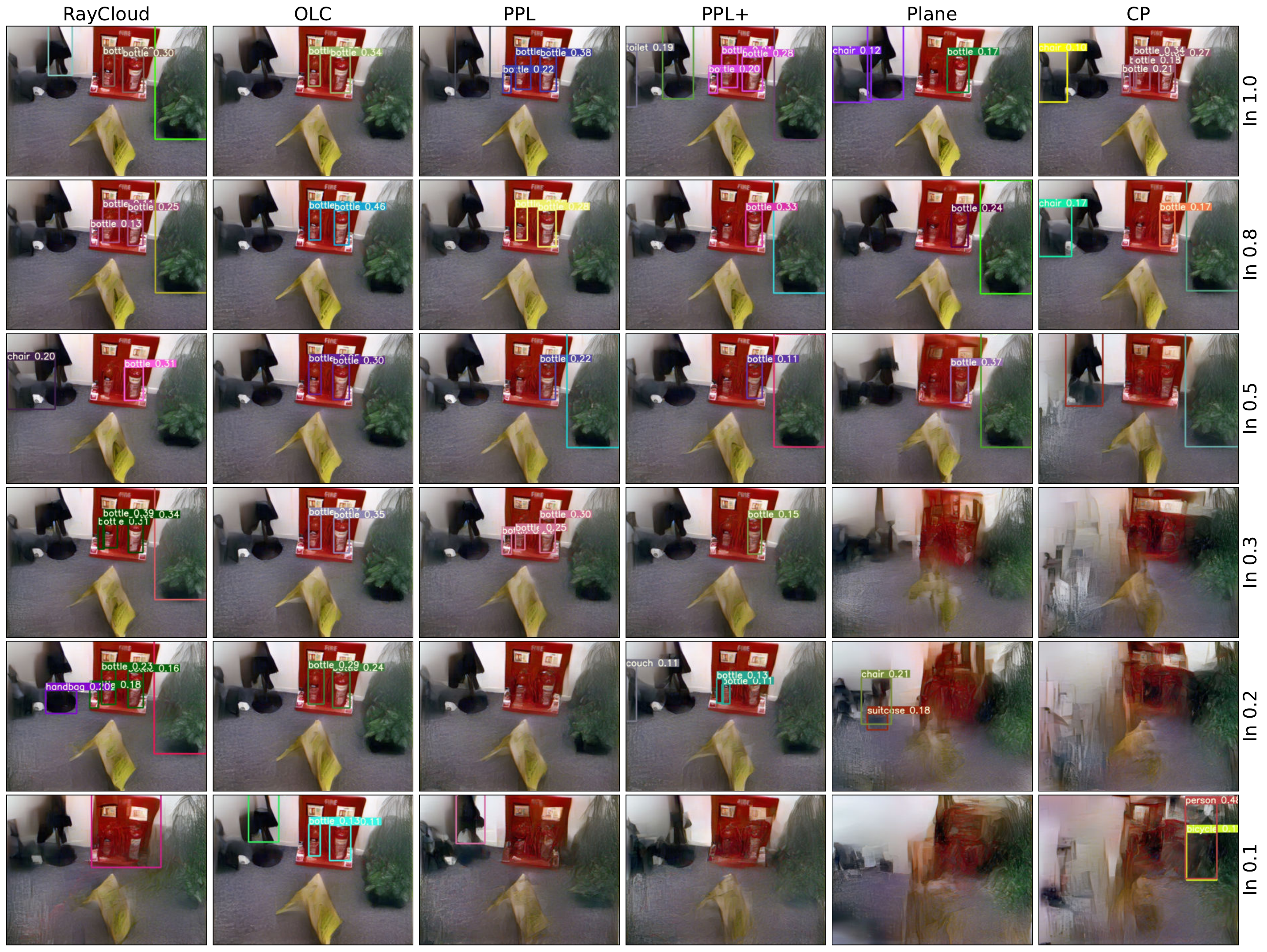}
  \caption{\textbf{Additional Qualitative Results - 7-scenes~\cite{shotton2013scene} dataset, scene \textit{Fire}}.
  Images inverted~\cite{pittaluga2019revealing} from the original points (`Baseline') and the points recovered from the 3D obfuscations from neighborhood information with various levels of inlier ratios (In.).
  Line obfuscations (OLC)~\cite{speciale2019privacy,speciale2019privacy2d}, Point-Pair-Lines PPL and PPL+~\cite{lee2023paired}, and ray clouds~\cite{moon2024efficient} are the most vulnerable to neighborhood-based attacks while Planes~\cite{geppert2022privacy} and Permutations~\cite{pan2023privacy} are more privacy preserving. The 3D points cloud is generated from SfM~\cite{schonberger2016structure} on SIFT~\cite{Lowe04IJCV} features.
  }
  \label{supp:quali_3D_fire}
\end{figure*}

As already observed in the main paper, the 3D line obfuscations OLC~\cite{speciale2019privacy}, PPL~\cite{lee2023paired}, PPL+~\cite{lee2023paired} and ray clouds~\cite{moon2024efficient} are the most susceptible to the point recovery, even when the neighborhood information is not reliable: more than 90\% of the points can be recovered with less than 10cm errors even when only 50\% of the nearest neighbor information is correct.
The image inversion from points recovered with only 10\% of inliers in the neighborhood still reveals the content of the original images, as can be seen in the last row of the Figures~\ref{supp:quali_3D_chess},~\ref{supp:quali_3D_redkitchen},~\ref{supp:quali_3D_office},~\ref{supp:quali_3D_fire}.
Out of the 3D line obfuscations, the most recent ray clouds appear to be the most privacy-preserving with the geometric accuracy dropping more as the inlier ratio of the neighborhoods decreases but the outline of the scene remains recognizable in the inverted images.
The point recovery also works on the plane~\cite{geppert2022privacy} and point-permutation~\cite{pan2023privacy} obfuscations but requires more reliable neighborhood information than for the 3D line obfuscations: the recovery is less accurate when the NN inlier ratio goes between 50\% and 30\%, which typically prevents meaningful image inversion.

\PAR{3D Point-Paired-Line Obfuscations: PPL and PPL+~\cite{lee2023paired}.}
The point-paired line obfuscations, PPL and PPL+~\cite{lee2023paired}, operate in 3D and transform the point cloud into a line cloud by generating lines joining random pairs of 3D points.
PPL+ is an extension of PPL that discourages lines to be formed between two points that lie on the same plane for two reasons: i) such lines could give hints on the scene structure, \eg, if the scene is a long corridor; ii) such lines are more vulnerable to density attacks~\cite{chelani2021privacy} as the distribution of line distances used to derive neighbors is more characteristic around each hidden points.

In our experiments, we observe that the performance of the proposed point recovery is equivalent between PPL and PPL+ as shown by the close geometric accuracies in Tables~\ref{supp:tab_geom_7scenes},~\ref{supp:tab_geom_12scenes},~\ref{supp:tab_geom_cambridge}.
These results are consistent with the original PPL paper~\cite{lee2023paired} where both PPL and PPL+ are recovered with similar errors by the density-based recovery~\cite{chelani2021privacy}.
One advantage of PPL over PPL+, though, is its faster runtime: PPL+ keeps drawing point-paired lines as long as the plane condition is not satisfied or until a certain number of iterations is reached.
When PPL can terminate in a matter of minutes on a small indoor point cloud typical of 12 scenes~\cite{valentin2016learning}, PPL+ can take several hours.

\begin{table*}[t]
\begin{center}
\begin{tabular}{ll*{12}{c}}
\toprule
& & \multicolumn{2}{c}{\textbf{OLC}~\cite{speciale2019privacy}}
 & \multicolumn{2}{c}{\textbf{PPL}~\cite{lee2023paired}}
 & \multicolumn{2}{c}{\textbf{PPL+}~\cite{lee2023paired}}
  & \multicolumn{2}{c}{\textbf{Rays}~\cite{moon2024efficient}}
 & \multicolumn{2}{c}{\textbf{Plane}~\cite{geppert2022privacy}} 
 & \multicolumn{2}{c}{\textbf{CP}~\cite{pan2023privacy}}\\ 
\cmidrule(r){3-4} \cmidrule(r){5-6} \cmidrule(r){7-8} \cmidrule(r){9-10} \cmidrule(r){11-12} \cmidrule(r){13-14}
\multirow{8}{*}{\rotatebox[origin=c]{90}{SIFT~\cite{Lowe04IJCV}}} & In. & 10cm & 25cm & 10cm & 25cm & 10cm & 25cm & 10cm & 25cm & 10cm & 25cm & 10cm & 25cm \\ 
\cmidrule(r){3-4} \cmidrule(r){5-6} \cmidrule(r){7-8} \cmidrule(r){9-10} \cmidrule(r){11-12} \cmidrule(r){13-14}
& 1.0  & 96.1 & 98.6 & 94.6 & 97.3 & 94.8 & 97.4 & 94.6 & 97.9 & 93.4 & 97.5 & 88.2 & 94.5\\
& 0.75 & 96.0 & 98.2 & 94.7 & 97.1 & 94.9 & 97.3 & 93.3 & 96.8 & 93.0 & 97.0 & 89.1 & 95.8\\ 
& 0.50 & 96.2 & 98.2 & 95.0 & 97.2 & 95.1 & 97.3 & 91.9 & 95.7 & 82.8 & 88.7 & 67.7 & 75.0\\ 
& 0.30 & 96.4 & 98.2 & 94.8 & 97.1 & 94.9 & 97.2 & 86.2 & 90.5 & 42.1 & 60.4 & 40.9 & 46.2\\ 
& 0.20 & 96.3 & 98.2 & 94.0 & 96.8 & 94.1 & 96.9 & 78.7 & 83.6 & 20.9 & 39.6 & 31.1 & 35.1\\ 
& 0.10 & 92.5 & 96.1 & 78.2 & 84.5 & 78.3 & 84.7 & 49.9 & 57.1 & 7.5 & 20.7 & 22.8 & 26.2\\
\bottomrule
\end{tabular}
\begin{tabular}{ll*{10}{c}}
& & \multicolumn{2}{c}{\textbf{OLC}~\cite{speciale2019privacy}}
 & \multicolumn{2}{c}{\textbf{PPL}~\cite{lee2023paired}}
  & \multicolumn{2}{c}{\textbf{Rays}~\cite{moon2024efficient}}
 & \multicolumn{2}{c}{\textbf{Plane}~\cite{geppert2022privacy}} 
 & \multicolumn{2}{c}{\textbf{CP}~\cite{pan2023privacy}}\\ 
\cmidrule(r){3-4} \cmidrule(r){5-6} \cmidrule(r){7-8} \cmidrule(r){9-10} \cmidrule(r){11-12}
\multirow{8}{*}{\rotatebox[origin=c]{90}{SuperPoint~\cite{detone2018superpoint}}} & In. & 10cm & 25cm & 10cm & 25cm & 10cm & 25cm & 10cm & 25cm & 10cm & 25cm \\ 
\cmidrule(r){3-4} \cmidrule(r){5-6} \cmidrule(r){7-8} \cmidrule(r){9-10} \cmidrule(r){11-12}
& 1.0  & 98.3 & 99.7 & 96.9 & 99.0 & 94.7 & 98.2 & 95.6 & 99.0 & 89.8 & 96.1 \\ 
& 0.75 & 98.2 & 99.6 & 97.2 & 99.0 & 93.3 & 97.2 & 95.1 & 98.8 & 90.4 & 97.6  \\ 
& 0.50 & 98.5 & 99.6 & 97.3 & 99.1 & 91.9 & 96.2 & 82.3 & 90.5 & 66.5 & 74.8 \\ 
& 0.30 & 98.6 & 99.6 & 96.6 & 98.8 & 85.8 & 91.1 & 40.8 & 63.4 & 39.8 & 45.8 \\ 
& 0.20 & 98.5 & 99.6 & 94.8 & 97.8 & 77.7 & 83.8 & 21.8 & 44.1 & 30.4 & 35.3 \\ 
& 0.10 & 93.5 & 97.0 & 72.7 & 80.5 & 47.3 & 55.9 & 9.6  & 28.3 & 21.7 & 26.0\\
\bottomrule
\end{tabular}
\caption{\textbf{Geometric accuracies $\uparrow$ of the 3D point recovery on the indoor 7-scenes~\cite{shotton2013scene} with SIFT~\cite{Lowe04IJCV} and SuperPoint~\cite{detone2018superpoint}}.
The point clouds are generated with Structure-from-Motion~\cite{schonberger2016structure}.
The performance of the point recovery is consistent between the 3D models generated from SuperPoint and SIFT~\cite{Lowe04IJCV} with variations in geometric accuracy in the order of a few percent, up to $~8$\% with the worst inlier ratio of 0.1.
This shows that the method is insensitive to the features used to generate the 3D model, which is not that surprising given that the optimization in the point recovery relies on the geometry only.
The line obfuscations OLC~\cite{speciale2019privacy}, PPL~\cite{lee2023paired}, PPL+~\cite{lee2023paired} and Ray clouds~\cite{moon2024efficient} are the most susceptible to the recovery, even when the neighborhood information is not reliable.
The point recovery also works on the plane~\cite{geppert2022privacy} and point-permutation~\cite{pan2023privacy} obfuscations but requires more reliable neighborhood information than for the previous obfuscation.
}
\label{supp:tab_geom_7scenes}
\end{center}
\end{table*}

\begin{table*}[t]
\begin{center}
\begin{tabular}{l*{12}{c}}
\toprule
& \multicolumn{2}{c}{\textbf{OLC}~\cite{speciale2019privacy}}
 & \multicolumn{2}{c}{\textbf{PPL}~\cite{lee2023paired}}
 & \multicolumn{2}{c}{\textbf{PPL+}~\cite{lee2023paired}}
  & \multicolumn{2}{c}{\textbf{Rays}~\cite{moon2024efficient}}
 & \multicolumn{2}{c}{\textbf{Plane}~\cite{geppert2022privacy}} 
 & \multicolumn{2}{c}{\textbf{CP}~\cite{pan2023privacy}}\\ 
\cmidrule(r){2-3} \cmidrule(r){4-5} \cmidrule(r){6-7} \cmidrule(r){8-9} \cmidrule(r){10-11} \cmidrule(r){12-13}
In. & 10cm & 25cm & 10cm & 25cm & 10cm & 25cm & 10cm & 25cm & 10cm & 25cm & 10cm & 25cm \\ 
\cmidrule(r){2-3} \cmidrule(r){4-5} \cmidrule(r){6-7} \cmidrule(r){8-9} \cmidrule(r){10-11} \cmidrule(r){12-13} 
1.0  & 99.2 & 99.8 & 98.8 & 99.6 & 98.4 & 99.1 & 97.7 & 99.2 & 99.0 & 99.7 & 92.3 & 96.2\\ 
0.75 & 99.3 & 99.8 & 98.8 & 99.6 & 98.9 & 99.6 & 95.0 & 97.7 & 97.4 & 98.2 & 94.2 & 98.9\\ 
0.50 & 99.3 & 99.8 & 98.6 & 99.6 & 98.8 & 99.6 & 93.3 & 97.3 & 79.3 & 84.2 & 72.7 & 79.7\\ 
0.30 & 99.4 & 99.8 & 97.8 & 99.5 & 98.2 & 99.5 & 91.5 & 96.5 & 38.5 & 54.0 & 39.6 & 44.2\\ 
0.20 & 99.3 & 99.8 & 96.0 & 98.7 & 96.8 & 99.0 & 88.4 & 94.0 & 20.0 & 36.1 & 27.6 & 31.0 \\ 
0.10 & 98.6 & 99.7 & 85.6 & 90.8 & 87.5 & 92.1 & 71.1 & 78.4 & 8.0  & 20.2 & 18.5 & 21.8 \\
\bottomrule
\end{tabular}
\caption{\textbf{Geometric accuracies $\uparrow$ of the 3D point recovery on the indoor 12-scenes~\cite{valentin2016learning} with SIFT~\cite{Lowe04IJCV}}. The conclusions are consistent with the results on the other indoor dataset 7-scenes~\cite{shotton2013scene} reported in Table~\ref{supp:tab_geom_7scenes}: the line obfuscations OLC~\cite{speciale2019privacy}, PPL~\cite{lee2023paired}, PPL+~\cite{lee2023paired} and Ray clouds~\cite{moon2024efficient} are the most susceptible to the recovery, even when the neighborhood information is not reliable (\eg 10\%), whereas the plane~\cite{geppert2022privacy} and point-permutation~\cite{pan2023privacy} are not recovered reliably as soon as the inlier ratio in the neighborhood information drops.}
\label{supp:tab_geom_12scenes}
\end{center}
\end{table*}

\begin{table*}[t]
\begin{center}
\begin{tabular}{l*{12}{c}}
\toprule
 & \multicolumn{2}{c}{\textbf{OLC}~\cite{speciale2019privacy}}
 & \multicolumn{2}{c}{\textbf{PPL}~\cite{lee2023paired}}
 & \multicolumn{2}{c}{\textbf{PPL+}~\cite{lee2023paired}}
  & \multicolumn{2}{c}{\textbf{Rays}~\cite{moon2024efficient}}
 & \multicolumn{2}{c}{\textbf{Plane}~\cite{geppert2022privacy}} 
 & \multicolumn{2}{c}{\textbf{Perm.}~\cite{pan2023privacy}}\\ 
\cmidrule(r){2-3} \cmidrule(r){4-5} \cmidrule(r){6-7} \cmidrule(r){8-9} \cmidrule(r){10-11} \cmidrule(r){12-13}
In. & 25cm & 50cm & 25cm & 50cm & 25cm & 50cm & 25cm & 50cm & 25cm & 50cm & 25cm & 50cm \\ 
\cmidrule(r){2-3} \cmidrule(r){4-5} \cmidrule(r){6-7} \cmidrule(r){8-9} \cmidrule(r){10-11} \cmidrule(r){12-13}
1.0  & 74.4 & 87.6 & 69.2 & 83.2 & 69.5 & 83.3 & 72.1 & 83.6 & 65.2 & 81.1 & 65.3 & 81.0\\ 
0.75 & 71.6 & 84.7 & 66.9 & 80.4 & 67.7 & 80.8 & 72.9 & 83.1 & 56.2 & 67.7 & 66.3 & 82.0\\ 
0.50 & 71.6 & 83.4 & 67.2 & 79.3 & 67.7 & 79.6 & 74.4 & 84.1 & 33.2 & 38.5 & 61.4 & 72.5\\ 
0.30 & 72.5 & 83.2 & 68.2 & 78.8 & 68.5 & 79.0 & 75.5 & 84.8 & 15.0 & 17.1 & 35.4 & 40.6\\ 
0.20 & 73.5 & 83.1 & 69.0 & 78.4 & 69.2 & 78.6 & 75.0 & 84.2 & 8.1 & 9.4 & 24.1 & 27.2\\ 
0.10 & 72.7 & 80.2 & 69.1 & 76.2 & 69.3 & 76.4 & 63.8 & 72.7 & 2.9 & 3.8 & 16.5 & 18.2\\
\bottomrule
\end{tabular}
\caption{\textbf{Geometric accuracies $\uparrow$ of the 3D point recovery on the outdoor Cambridge~\cite{kendall2015posenet} dataset with SIFT~\cite{Lowe04IJCV}}. The same trend is observed outdoors as it is indoors, \ie, the line obfuscations OLC~\cite{speciale2019privacy}, PPL~\cite{lee2023paired}, PPL+~\cite{lee2023paired} and Ray clouds~\cite{moon2024efficient} are the most susceptible to the recovery, even when the neighborhood information is not reliable (\eg 10\%), whereas the plane~\cite{geppert2022privacy} and point-permutation~\cite{pan2023privacy} are not recovered reliably when the inlier ratio drops too low.
Although the geometric accuracy values are lower than for indoor and measured at higher error thresholds, the images inverted from the recovered points remain meaningful as shown in Figures~\ref{supp:quali_3D_shopfacade},~\ref{supp:quali_3D_kingscollege},~\ref{supp:quali_3D_oldhospital},~\ref{supp:quali_3D_stmaryschurch}.}
\label{supp:tab_geom_cambridge}
\end{center}
\end{table*}

\PAR{Perceptual Evaluation of other 3D obfuscations.}
In the main paper, we reported only SSIM for the sake of clarity since the three metrics exhibit the same trend over all obfuscations and inlier ratios.
For the sake of completeness, we additionally report the SSIM and PSNR values for all 3D obfuscations on 7-scenes~\cite{shotton2013scene} (Table~\ref{supp:tab_perc_7scenes}), 12-scenes~\cite{valentin2016learning} (Table~\ref{supp:tab_perc_12scenes}), and Cambridge~\cite{kendall2015posenet} (Table~\ref{supp:tab_perc_cambridge}).
To keep the table readable, we report values only for PPL~\cite{lee2023paired} as the PPL+~\cite{lee2023paired} perceptual metrics are either equal or within 0.01 difference, which is negligible.

Similarly to the geometric evaluation, the recovery from the line obfuscations OLC~\cite{speciale2019privacy}, PPL~\cite{lee2023paired} and ray clouds~\cite{moon2024efficient} is stable across the inlier ratio of the neighborhood information whereas the recovery from the plane~\cite{geppert2022privacy} and the point permutation~\cite{pan2023privacy} is more sensitive to incorrect neighbors between 50\% and 30\% inlier ratios.

\begin{table*}[t]
\begin{center}
\begin{tabular}{l*{15}{c}}
\toprule
 & \multicolumn{5}{c}{\textbf{LPIPS$\downarrow$} Baseline: \textbf{0.52}}
 & \multicolumn{5}{c}{\textbf{SSIM$\uparrow$} Baseline: \textbf{0.58}} 
 & \multicolumn{5}{c}{\textbf{PSNR$\uparrow$} Baseline: \textbf{16.01}}\\ 
\cmidrule(r){2-6} \cmidrule(r){7-11} \cmidrule(r){12-16} 
In. & OLC & PPL & Ray & Plane & CP 
& OLC & PPL & Ray & Plane & CP 
& OLC & PPL & Ray & Plane & CP\\ 
\cmidrule(r){2-6} \cmidrule(r){7-11} \cmidrule(r){12-16} 
1.0  & 0.53 & 0.53 & 0.53 & 0.55 & 0.55 & 0.58 & 0.57 & 0.57 & 0.55 & 0.56 & 15.9 & 15.8 & 15.8 & 15.5 & 15.6\\ 
0.75 & 0.53 & 0.54 & 0.54 & 0.56 & 0.55 & 0.57 & 0.57 & 0.57 & 0.54 & 0.55 & 15.9 & 15.7 & 15.8 & 15.3 & 15.5\\ 
0.50 & 0.53 & 0.55 & 0.54 & 0.60 & 0.59 & 0.57 & 0.56 & 0.56 & 0.49 & 0.51 & 15.8 & 15.6 & 15.7 & 14.4 & 14.7\\
0.30 & 0.54 & 0.55 & 0.55 & 0.64 & 0.63 & 0.57 & 0.55 & 0.56 & 0.44 & 0.45 & 15.8 & 15.4 & 15.5 & 13.0 & 13.1\\ 
0.20 & 0.54 & 0.56 & 0.56 & 0.66 & 0.65 & 0.57 & 0.54 & 0.54 & 0.43 & 0.43 & 15.8 & 15.2 & 15.3 & 12.1 & 12.6\\
0.10 & 0.54 & 0.59 & 0.60 & 0.68 & 0.66 & 0.56 & 0.51 & 0.50 & 0.42 & 0.41 & 15.7 & 14.4 & 14.3 & 11.5 & 12.2\\ 
\bottomrule
\end{tabular}
\caption{\textbf{Perceptual metrics of the 3D point recovery on the indoor 7-scenes~\cite{shotton2013scene} with SIFT~\cite{Lowe04IJCV}}: the metrics assess how close the image inverted~\cite{pittaluga2019revealing} from the points recovered from the obfuscations is to the image inverted~\cite{pittaluga2019revealing} from the original points.
As for the geometric evaluation, the recovery from the line obfuscations OLC~\cite{speciale2019privacy}, PPL~\cite{lee2023paired} and Ray cloud~\cite{moon2024efficient} is stable across the inlier ratio of the neighborhood information whereas the recovery from the plane~\cite{geppert2022privacy} and the point permutation~\cite{pan2023privacy} is more sensitive to incorrect neighbors.
}
\label{supp:tab_perc_7scenes}
\end{center}
\end{table*}

\begin{table*}[t]
\begin{center}
\begin{tabular}{l*{15}{c}}
\toprule
 & \multicolumn{5}{c}{\textbf{LPIPS$\downarrow$} Baseline: \textbf{0.52}}
 & \multicolumn{5}{c}{\textbf{SSIM$\uparrow$} Baseline: \textbf{0.58}} 
 & \multicolumn{5}{c}{\textbf{PSNR$\uparrow$} Baseline: \textbf{16.01}}\\ 
\cmidrule(r){2-6} \cmidrule(r){7-11} \cmidrule(r){12-16} 
In. & OLC & PPL & Ray & Plane & CP 
& OLC & PPL & Ray & Plane & CP 
& OLC & PPL & Ray & Plane & CP\\ 
\cmidrule(r){2-6} \cmidrule(r){7-11} \cmidrule(r){12-16} 
1.0  & 0.56 & 0.57 & 0.57 & 0.58 & 0.58 & 0.49 & 0.49 & 0.48 & 0.48 & 0.47 & 14.6 & 14.5 & 14.3 & 14.3 & 14.3\\ 
0.75 & 0.57 & 0.57 & 0.58 & 0.60 & 0.58 & 0.49 & 0.48 & 0.48 & 0.44 & 0.47 & 14.6 & 14.5 & 14.3 & 13.9 & 14.2\\ 
0.50 & 0.57 & 0.58 & 0.58 & 0.64 & 0.61 & 0.49 & 0.48 & 0.47 & 0.40 & 0.43 & 14.5 & 14.4 & 14.3 & 12.9 & 13.3\\ 
0.30 & 0.57 & 0.58 & 0.58 & 0.66 & 0.65 & 0.49 & 0.47 & 0.47 & 0.36 & 0.38 & 14.5 & 14.3 & 14.2 & 11.8 & 12.1\\ 
0.20 & 0.57 & 0.59 & 0.58 & 0.67 & 0.66 & 0.49 & 0.47 & 0.47 & 0.35 & 0.36 & 14.5 & 14.2 & 14.2 & 11.3 & 11.6\\ 
0.10 & 0.57 & 0.60 & 0.60 & 0.69 & 0.67 & 0.49 & 0.45 & 0.45 & 0.34 & 0.34 & 14.5 & 13.8 & 13.8 & 10.8 & 11.2\\ 
\bottomrule
\end{tabular}
\caption{\textbf{Perceptual metrics of the 3D point recovery on the indoor 12-scenes~\cite{valentin2016learning}  with SIFT~\cite{Lowe04IJCV}}: the metrics assess how close the image inverted~\cite{pittaluga2019revealing} from the points recovered from the obfuscations is to the image inverted~\cite{pittaluga2019revealing} from the original points.
As for the geometric evaluation, the recovery from the line obfuscations OLC~\cite{speciale2019privacy}, PPL~\cite{lee2023paired}  and Ray cloud~\cite{moon2024efficient} is stable across the inlier ratio of the neighborhood information whereas the recovery from the plane~\cite{geppert2022privacy} and the point permutation~\cite{pan2023privacy} is more sensitive to incorrect neighbors.
}
\label{supp:tab_perc_12scenes}
\end{center}
\end{table*}

\begin{table*}[t]
\begin{center}
\begin{tabular}{l*{15}{c}}
\toprule
 & \multicolumn{5}{c}{\textbf{LPIPS$\downarrow$} Baseline: \textbf{0.52}}
 & \multicolumn{5}{c}{\textbf{SSIM$\uparrow$} Baseline: \textbf{0.58}} 
 & \multicolumn{5}{c}{\textbf{PSNR$\uparrow$} Baseline: \textbf{16.01}}\\ 
\cmidrule(r){2-6} \cmidrule(r){7-11} \cmidrule(r){12-16} 
In. & OLC & PPL & Ray & Plane & CP 
& OLC & PPL & Ray & Plane & CP 
& OLC & PPL & Ray & Plane & CP\\ 
\cmidrule(r){2-6} \cmidrule(r){7-11} \cmidrule(r){12-16} 
1.0  & 0.64 & 0.64 & 0.63 & 0.64 & 0.64 & 0.37 & 0.36 & 0.37 & 0.36 & 0.36 & 12.8 & 12.7 & 12.8 & 12.7 & 12.7\\ 
0.75 & 0.64 & 0.64 & 0.63 & 0.66 & 0.64 & 0.36 & 0.36 & 0.37 & 0.34 & 0.36 & 12.7 & 12.6 & 12.8 & 12.2 & 12.7\\ 
0.50 & 0.64 & 0.64 & 0.63 & 0.67 & 0.66 & 0.36 & 0.36 & 0.37 & 0.32 & 0.34 & 12.6 & 12.6 & 12.8 & 11.6 & 12.2\\ 
0.30 & 0.64 & 0.64 & 0.63 & 0.69 & 0.69 & 0.36 & 0.36 & 0.37 & 0.31 & 0.29 & 12.6 & 12.5 & 12.8 & 10.9 & 11.2 \\ 
0.20 & 0.64 & 0.65 & 0.64 & 0.70 & 0.70 & 0.36 & 0.36 & 0.37 & 0.31 & 0.27 & 12.5 & 12.5 & 12.7 & 10.5 & 10.8\\ 
0.10 & 0.65 & 0.65 & 0.65 & 0.71 & 0.70 & 0.34 & 0.35 & 0.35 & 0.30 & 0.26 & 12.3 & 12.4 & 12.4 & 10.3 & 10.6\\ 
\bottomrule
\end{tabular}
\caption{\textbf{Perceptual metrics of the 3D point recovery on the outdoor Cambridge~\cite{kendall2015posenet}  with SIFT~\cite{Lowe04IJCV}}: the metrics assess how close the image inverted~\cite{pittaluga2019revealing} from the points recovered from the obfuscations is to the image inverted~\cite{pittaluga2019revealing} from the original points.
Similarly to the geometric evaluation, the recovery from the line obfuscations OLC~\cite{speciale2019privacy}, PPL~\cite{lee2023paired}, and Ray clouds~\cite{moon2024efficient} is stable across the inlier ratio of the neighborhood information whereas the recovery from the plane~\cite{geppert2022privacy} and the point permutation~\cite{pan2023privacy} is more sensitive to incorrect neighbors.
}
\label{supp:tab_perc_cambridge}
\end{center}
\end{table*}

\PAR{Comparison to other 3D line recoveries.}
We compare the proposed point recovery to the existing density-based recoveries in~\cite{chelani2021privacy} and~\cite{lee2023paired} that operate on 3D lines only (Table~\ref{supp:baseline_olc_ppl}).
These methods estimate the neighborhood of a given 3D line based on the density of all lines in the cloud and the original point is approximated with the position of highest density along the line.
We observe that our method largely outperforms those baselines even with as little as 20\% inlier ratio in the neighborhood information necessary for our recovery.

However, we note that the results for~\cite{lee2023paired} computed with the author's public release seem subpar to the results reported in the paper so this comparison should be taken as an indicative result only.
We believe that this discrepancy in the results is not due to a technical issue in the method or the code of~\cite{lee2023paired} but rather the difference in input data: the point clouds we generated and the points clouds of~\cite{lee2023paired} are most likely different because of variations in the Structure-from-Motion~\cite{schonberger2016structure}, \eg, because of differences in the parameters or the randomness of the robust geometric estimation.
To reduce the potential discrepancy in the input data and for this experiment only, we use the point clouds used in~\cite{chelani2021privacy} to run this evaluation instead of the point clouds we generated for the rest of the paper.
However, discrepancies between the input data used in~\cite{chelani2021privacy} and ~\cite{lee2023paired} remain and this is why these results should be taken as indicative results only.

\begin{table}[t]
\fontsize{9}{6.2}\selectfont
\begin{center}
\begin{tabular}{l*{2}{c}}
\toprule
Recovery       &     \textbf{OLC 3D lines}             & \textbf{PPL 3D lines}         \\
\cmidrule(r){1-1} \cmidrule(r){2-2} \cmidrule(r){3-3}
                              & 5 / 10 / 25 cm            & 5 / 10 / 25 cm \\
\cmidrule(r){2-2} \cmidrule(r){3-3}
OLC Rec.~\cite{chelani2021privacy} & 67.5 / 75.8 / 84.0      &         $-$     \\
PPL Rec.~\cite{lee2023paired}      & $-$                     &  34.85 / 48.71 / 63.16 \\
\cmidrule(r){2-2} \cmidrule(r){3-3}
\textbf{Ours} 50\% In.                      &  94.6 / 99.3 / 99.9    &  89.7 / 98.2 / 99.5 \\
\textbf{Ours} 20\% In.                      &  91.7 / 99.1 / 99.8    &  82.1 / 93.9 / 97.4 \\
\bottomrule
\end{tabular}
\caption{\textbf{Comparison against 3D line recovery baselines}. Geometric accuracy $\uparrow$ of the recovery from 3D obfuscations against baseline methods~\cite{chelani2021privacy,lee2023paired} on 12-scenes.
The recoveries are run on the same 12scenes~\cite{valentin2016learning} point clouds as in~\cite{chelani2021privacy}, which differ from the point clouds used in the rest of the paper that we generated ourselves with COLMAP~\cite{schonberger2016structure} or from the points clouds from~\cite{lee2023paired}.
}
\label{supp:baseline_olc_ppl}
\end{center}
\end{table}

\begin{table}[t]
\fontsize{8.7}{6.3}\selectfont
\begin{center}
\begin{tabular}{l*{5}{c}}
\toprule
\textbf{Num. Pts}  & \textbf{In 1.0}  & \textbf{In 0.50}  & \textbf{In 0.30}  & \textbf{In 0.20}  & \textbf{In 0.10}  \\
\midrule
700K            &   0:41   &    0:51    & 1:30       &   4:00 & 18:00 \\
300K            &   0:19   &    0:24    & 0:45       &   1:20 & 9:00 \\
100K            &   0:08   &    0:10    & 0:19       &   0:46 &  3:40 \\
\bottomrule
\end{tabular}
\caption{\textbf{Indicative runtime} as a function of the number of 3D points (Num.Pts) and inlier ratios (In.) for the recovery from the PPL~\cite{lee2023paired} obfuscation with 50 neighbors.
The 3D points cloud is generated from SfM~\cite{schonberger2016structure} on SIFT~\cite{Lowe04IJCV} features.
X:Y indicates that the runtime takes X minutes and Y seconds.
The theoretical number of RANSAC~\cite{fischler1981random} in the optimization is inversely proportional to the inlier ratio, hence the longer runtimes as the inlier ratio decreases.
Still, the runtime remains small enough that the point recovery is practical for an attacker.
The recovery runs on a single AMD EPYC CPU with 64 cores.
}
\label{supp:runtime_3d}
\end{center}
\end{table}

\PAR{Influence of the features on the point recovery in 3D.}
We assess whether the performance of the point recovery depends on the type of features extracted from the images and used for the Structure-from-Motion~\cite{schonberger2016structure} that generates the 3D point cloud.
We compare the hand-crafted SIFT~\cite{Lowe04IJCV} and the deep-learning-based SuperPoint~\cite{detone2018superpoint} and report the geometric accuracy on 7-scenes~\cite{shotton2013scene} in~\cref{supp:tab_geom_7scenes}.
The performance of the point recovery is consistent between the 3D models generated from SuperPoint~\cite{detone2018superpoint} and SIFT~\cite{Lowe04IJCV} with variations in geometric accuracy in the order of a few percent.
This shows that the method is insensitive to the features used to generate the 3D model, which is not that surprising given that the optimization in the point recovery relies on the geometry only.

\begin{figure*}
  \centering
    \includegraphics[width=0.5\linewidth,height=\textheight,keepaspectratio]{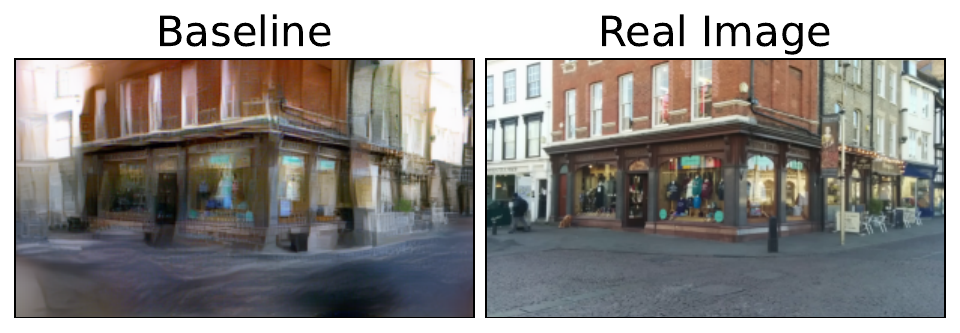}
    \includegraphics[width=\linewidth,height=\textheight,keepaspectratio]
    {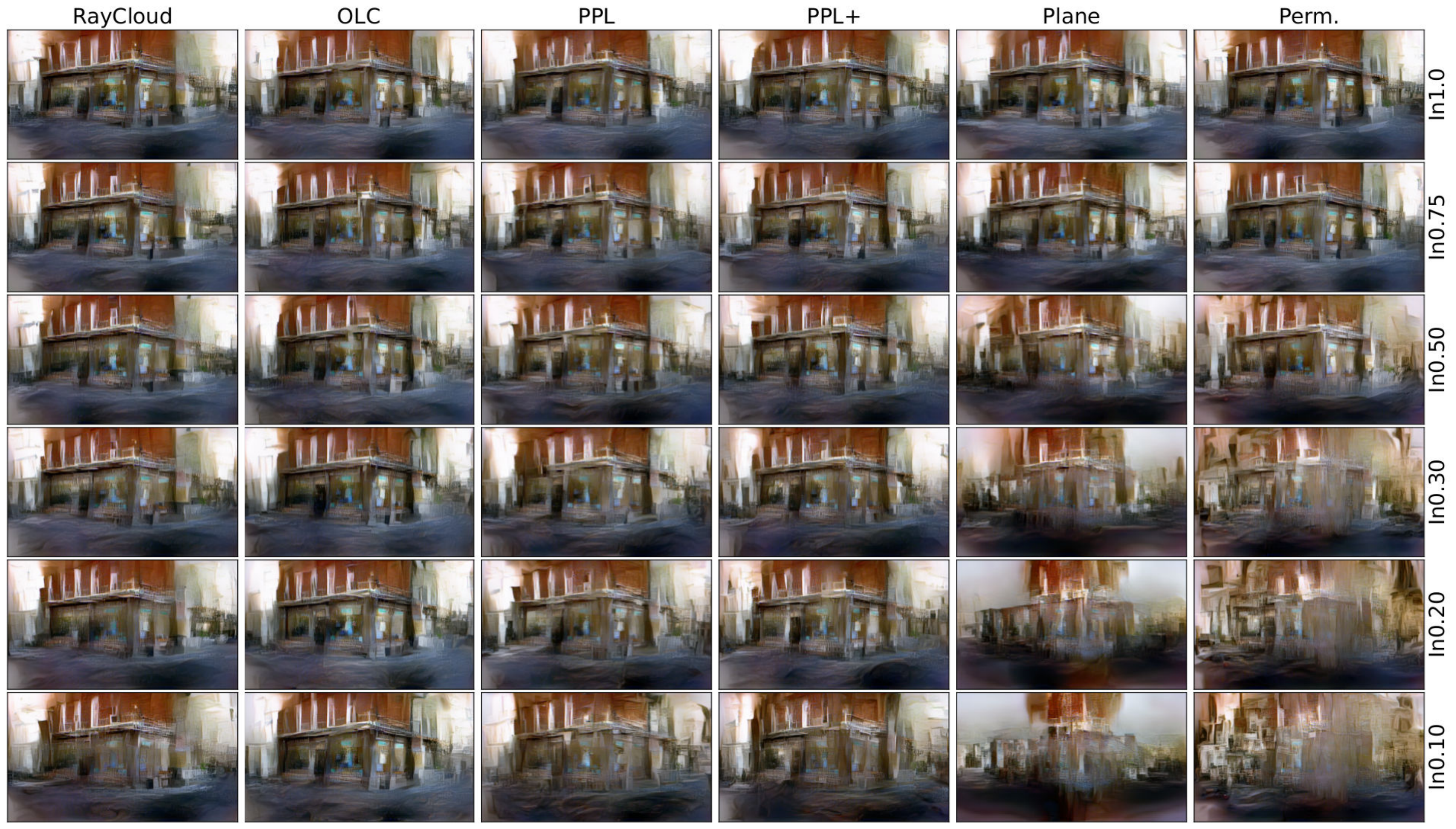
    }
  \caption{\textbf{Additional Qualitative Results - Cambridge~\cite{kendall2015posenet} dataset, scene \textit{Shop Facade}}.
  Images inverted~\cite{pittaluga2019revealing} from the original points (`Baseline') and the points recovered from the 3D obfuscations from neighborhood information with various levels of inlier ratios (In.).
  Line obfuscations (OLC)~\cite{speciale2019privacy,speciale2019privacy2d}, Point-Pair-Lines PPL and PPL+~\cite{lee2023paired} and ray clouds~\cite{moon2024efficient} are the most vulnerable to neighborhood-based attacks while Planes~\cite{geppert2022privacy} and Permutations~\cite{pan2023privacy} are more privacy preserving. The 3D points cloud is generated from SfM~\cite{schonberger2016structure} on SIFT~\cite{Lowe04IJCV} features.
  }
  \label{supp:quali_3D_shopfacade}
\end{figure*}

\begin{figure*}
  \centering
    \includegraphics[width=0.5\linewidth,height=\textheight,keepaspectratio]{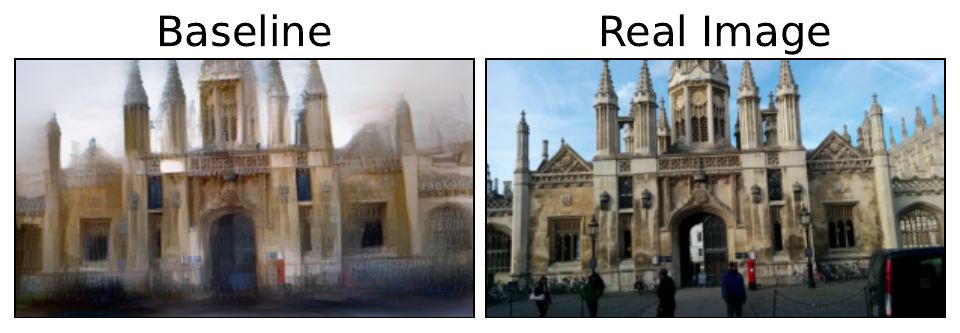}
    \includegraphics[width=\linewidth,height=\textheight,keepaspectratio]
    {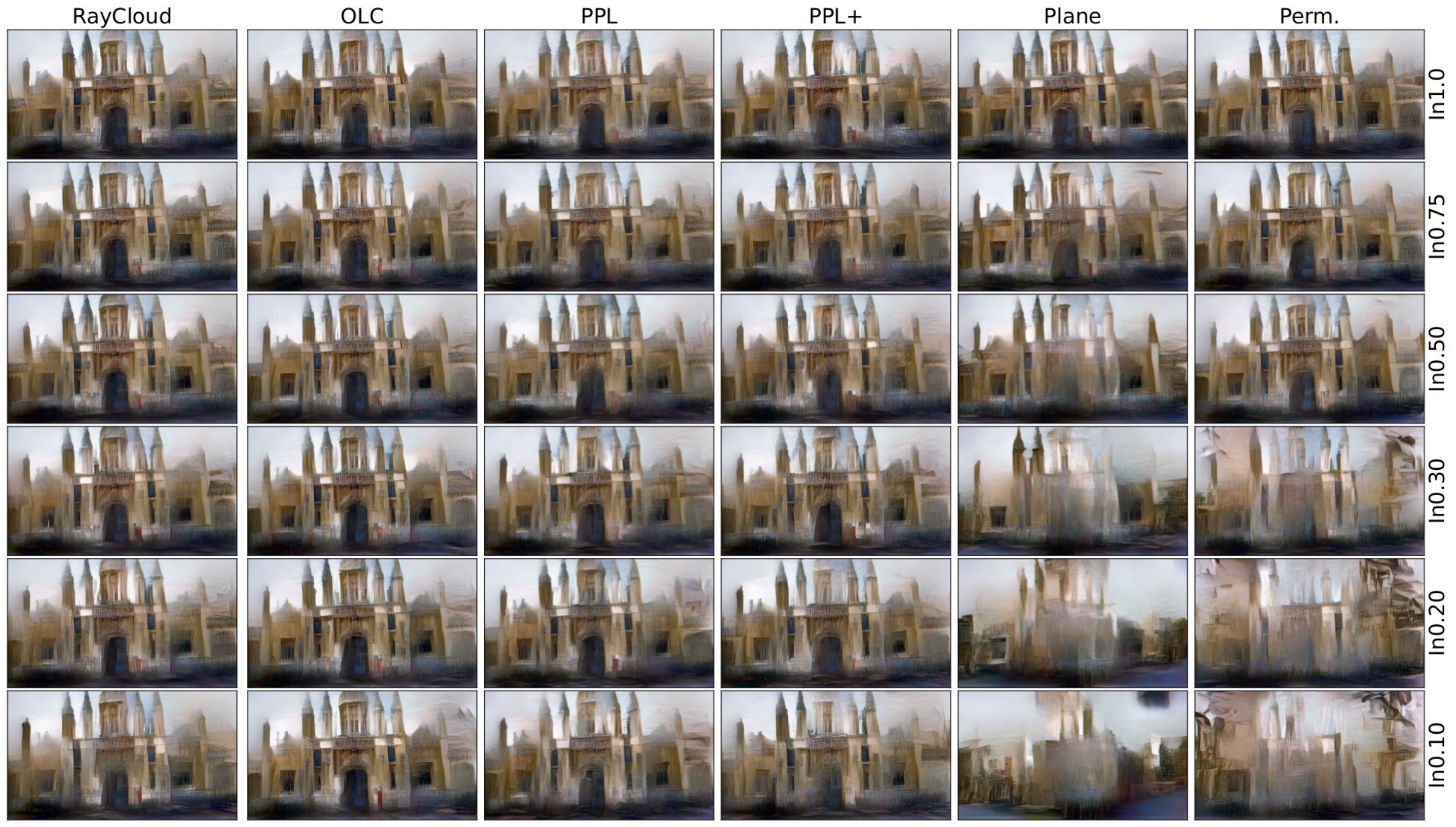
    }
  \caption{\textbf{Additional Qualitative Results - Cambridge~\cite{kendall2015posenet} dataset, scene \textit{King's College}}.
  Images inverted~\cite{pittaluga2019revealing} from the original points (`Baseline') and the points recovered from the 3D obfuscations from neighborhood information with various levels of inlier ratios (In.).
  Line obfuscations (OLC)~\cite{speciale2019privacy,speciale2019privacy2d}, Point-Pair-Lines PPL and PPL+~\cite{lee2023paired}, and ray clouds~\cite{moon2024efficient} are the most vulnerable to neighborhood-based attacks while Planes~\cite{geppert2022privacy} and Permutations~\cite{pan2023privacy} are more privacy preserving. The 3D points cloud is generated from SfM~\cite{schonberger2016structure} on SIFT~\cite{Lowe04IJCV} features.
  }
    \label{supp:quali_3D_kingscollege}
\end{figure*}

\begin{figure*}
  \centering
    \includegraphics[width=0.5\linewidth,height=\textheight,keepaspectratio]{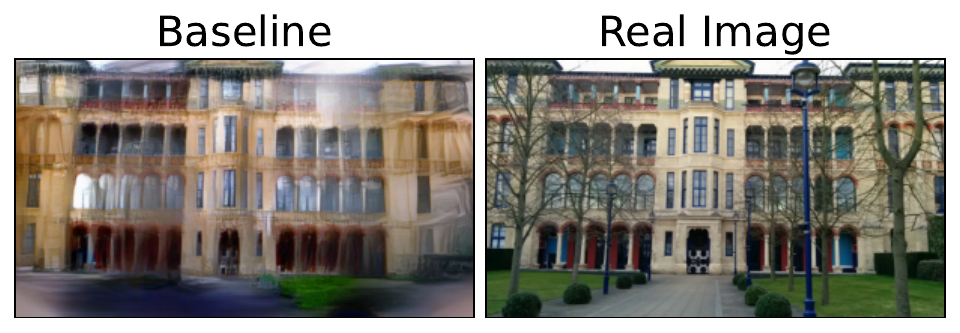
    }
    \includegraphics[width=\linewidth,height=\textheight,keepaspectratio]
    {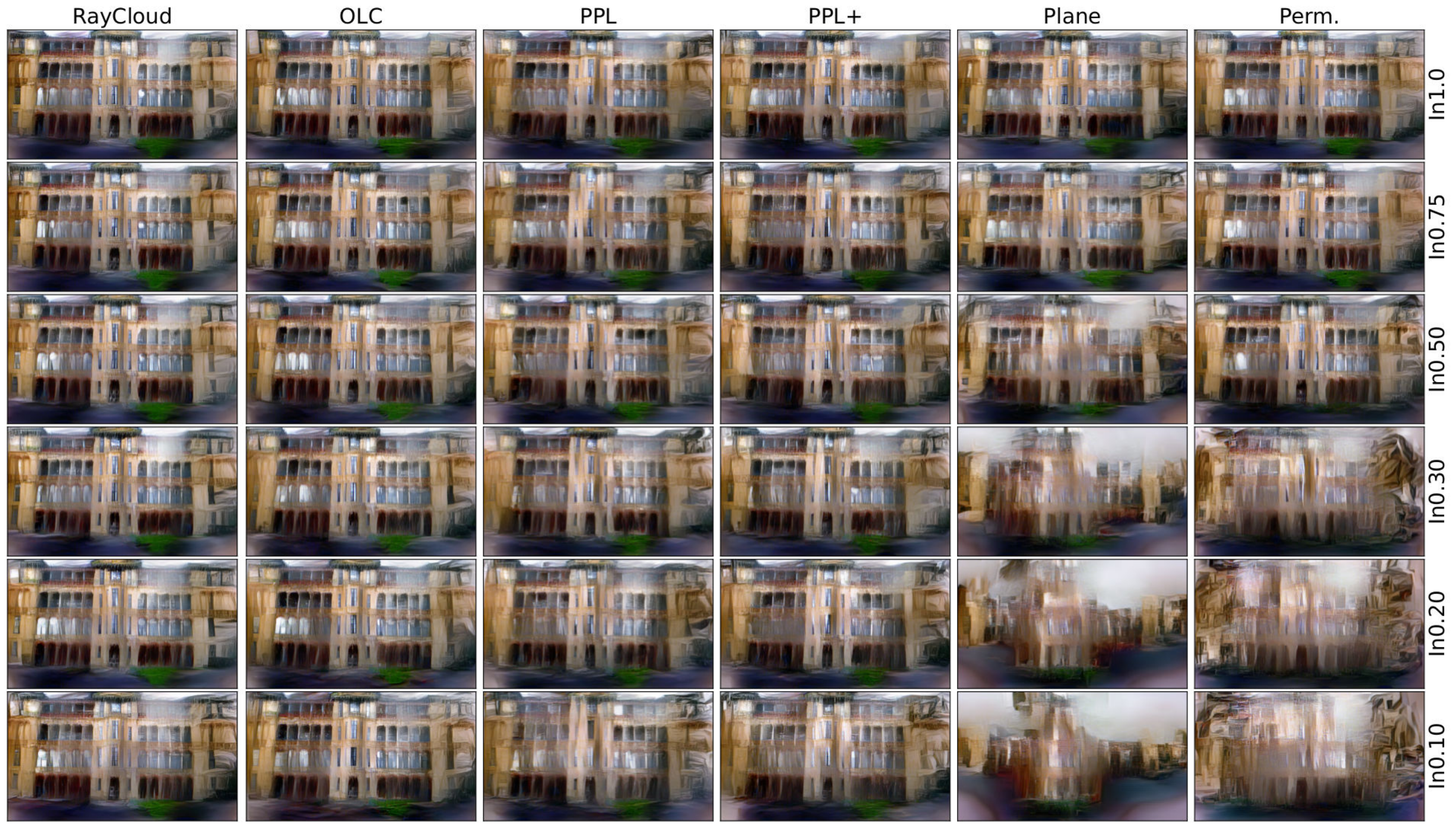
    }
  \caption{\textbf{Additional Qualitative Results - Cambridge~\cite{kendall2015posenet} dataset, scene \textit{ Old Hospital}}.
  Images inverted~\cite{pittaluga2019revealing} from the original points (`Baseline') and the points recovered from the 3D obfuscations from neighborhood information with various levels of inlier ratios (In.).
  Line obfuscations (OLC)~\cite{speciale2019privacy,speciale2019privacy2d}, Point-Pair-Lines PPL and PPL+~\cite{lee2023paired}, and ray clouds~\cite{moon2024efficient} are the most vulnerable to neighborhood-based attacks while Planes~\cite{geppert2022privacy} and Permutations~\cite{pan2023privacy} are more privacy preserving. The 3D points cloud is generated from SfM~\cite{schonberger2016structure} on SIFT~\cite{Lowe04IJCV} features.
  }
      \label{supp:quali_3D_oldhospital}
\end{figure*}

\begin{figure*}
  \centering
    \includegraphics[width=0.5\linewidth,height=\textheight,keepaspectratio]{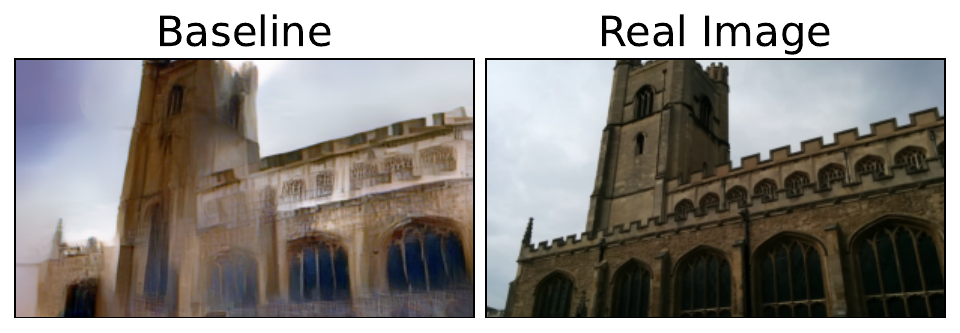
    }
    \includegraphics[width=\linewidth,height=\textheight,keepaspectratio]
    {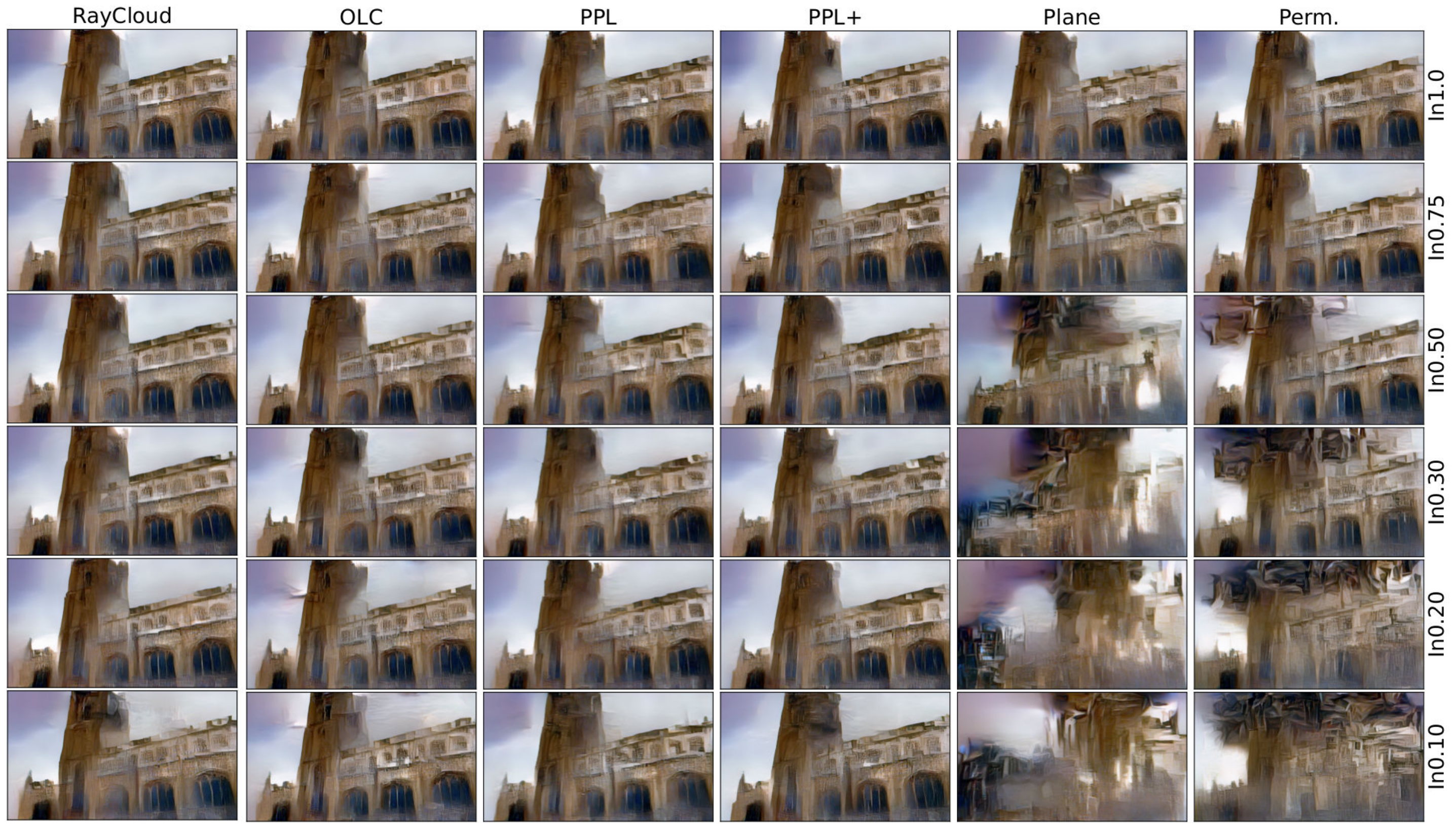
    }
  \caption{\textbf{Additional Qualitative Results - Cambridge~\cite{kendall2015posenet} dataset, scene \textit{St. Mary's Church}}.
  Images inverted~\cite{pittaluga2019revealing} from the original points (`Baseline') and the points recovered from the 3D obfuscations from neighborhood information with various levels of inlier ratios (In.).
  Line obfuscations (OLC)~\cite{speciale2019privacy,speciale2019privacy2d}, Point-Pair-Lines PPL and PPL+~\cite{lee2023paired}, and ray clouds~\cite{moon2024efficient} are the most vulnerable to neighborhood-based attacks while Planes~\cite{geppert2022privacy} and Permutations~\cite{pan2023privacy} are more privacy preserving. The 3D points cloud is generated from SfM~\cite{schonberger2016structure} on SIFT~\cite{Lowe04IJCV} features.
  The 3D points cloud is generated from SfM~\cite{schonberger2016structure} on SIFT~\cite{Lowe04IJCV} features.
  }
\label{supp:quali_3D_stmaryschurch}
\end{figure*}

\PAR{Qualitative Results.}
Further examples of images inverted from the points recovered from various obfuscations are shown in
Figures~\ref{supp:quali_3D_chess},~\ref{supp:quali_3D_redkitchen},~\ref{supp:quali_3D_office},~\ref{supp:quali_3D_fire} on 7scenes~\cite{shotton2013scene},
and in Figures~\ref{supp:quali_3D_shopfacade},~\ref{supp:quali_3D_kingscollege},~\ref{supp:quali_3D_oldhospital},~\ref{supp:quali_3D_stmaryschurch} on Cambridge~\cite{kendall2015posenet}.

\PAR{Detected content.}
In addition to the previous perceptual evaluation, we measure the recovered information at the finer level of objects and adopt the SegLoc's evaluation~\cite{pietrantoni2023segloc}.
An off-the-shelf object detector, YoloV7~\cite{wang2022yolov7}, runs on both %
the images inverted from the original points and the recovered points.
The discrepancy between the two sets of detections is a relevant proxy to measure how much content is recovered. 
We report the standard detection metric in Table~\ref{supp:tab_detection} where the detections on the real images are used as ground-truth and the detections on the images inverted from the original points clouds are the baseline.
For the sake of clarity, we only report here the recall of the detection for it indicates the amount of objects discovered by the attack, which is more relevant than the precision at which the object is discovered. 
These values are indicative only as when we appraise the inverted images visually, it often occurs that the inverted images is decipherable by the human eye but the detection fails to identify the objects because of the domain shift and the noise of the image.
Hence, the detection performance tends to over-estimate the privacy of the evaluated representations.

\begin{table*}[t]
\begin{center}
\begin{tabular}{l*{7}{c}}
\toprule
& \multicolumn{4}{c}{\textbf{3D}} & \multicolumn{3}{c}{\textbf{2D}}\\ 
\cmidrule(r){2-5} \cmidrule(r){6-8}
\textbf{Object} & Baseline & PPL         & Plane      & Perm.        & Baseline & Line & Perm. \\ 
\cmidrule(r){2-2} \cmidrule(r){3-5} \cmidrule(r){6-6} \cmidrule(r){7-8}
\textbf{TV}       &   19.5   & 11.4 / 15.3 & 5.6 / 13.5 & 7.1 / 13.6 &   16.0    &  7.0 / 5.8   & 2.4 / 5.0 \\
\textbf{Backpack} &   21.1   & 11.7 / 17.5 & 3.6 / 5.1  & 8.8 / 8.0  &   14.6    &  1.5 / 0.7   & 0   / 1.4  \\
\textbf{Plant}    &   23.0   & 10.5 / 25.4 & 5.7 / 25.4 & 5.7 / 22.9 &   34.4    &  17.2 / 13.1 & 9.8 / 16.4\\
\bottomrule
\end{tabular}
\caption{\textbf{Private content detected} on the images inverted from the recovered point.
The original points are derived from SIFT~\cite{Lowe04IJCV}.
The detection~\cite{wang2022yolov7} on the original images serves as ground-truth and the baseline indicates the performance of the detection on the images inverted from the original points.
We report the detection recall $\uparrow$ on points recovered from neighborhood information at inlier ratios (0.50 / 1.0).
Even though the recall of the images inverted from obfuscations is lower than the baseline, we observe that this evaluation under-estimate the amount of private content that is revealed.
This is because the off-the-shelf detector is typically subpar on the inverted images: it fails to detect objects that the human eye can still perceive, which is usually because of the distribution shift in the image pixels or because of the noise in the image.
}
\label{supp:tab_detection}
\end{center}
\end{table*}

\section{Implementation Details}
\label{supp:impl_details}

\PAR{Geometric Recovery and Runtime.}
The point recovery runs within a reasonable amount of time: the minimization is implemented using the open-source Ceres~\cite{Agarwal_Ceres_Solver_2022} optimization library and runs in parallel on a single CPU.
The runtime is a function of the number of points in the point cloud or the image, the inlier ratio, the neighborhood size, and the maximum number of RANSAC~\cite{fischler1981random} iterations: the more points and the larger the neighborhood, the more time the computation takes.
In parallel, the higher the inlier ratio, the lower the runtime as the optimal number of RANSAC~\cite{fischler1981random} iterations is inversely proportional to the inlier ratio.
For example, the biggest point cloud in the experiments has 700K points (12-scenes-office1-gates-381~\cite{valentin2016learning}).
In a setup with 100 neighbors and an upper bound on the number of RANSAC~\cite{fischler1981random} iterations set to 10K, the runtime varies between 1 minute 30 s when there are no outliers in the neighborhood up to 4 minutes for inlier ratios between 75\% and 20\%, on a single AMD EPYC CPU with 64 cores.
Table~\ref{supp:runtime_3d} gives more runtime examples as a function of the point cloud size and inlier ratios.

\PAR{Point Initialization.}
We observe that the point recovery is insensitive to the point initialization and use the following heuristics in the paper.

For 3D lines~\cite{speciale2019privacy,lee2023paired,moon2024efficient} and 3D lines made from 3D permutation~\cite{pan2023privacy}, the 3D points to recover are initialized as the projection of a 3D point "anchor" onto the lines.
The 3D anchor is defined as follows: the 3D lines are projected onto a plane.
We sample a set of intersections between the resulting 2D lines and compute their 2D centroid, which we use as the anchor.
In the paper, we use the plane $z=0$ so the centroid has the form $(x,y,0)$ and randomly sample 10K intersection points.
Note that the choice for the plane $z=0$ does not necessarily correspond to the ground-plane as the coordinate frames of the scenes are chosen arbitrarily by the authors of the datasets.

The initialization in 2D follows the same steps except that the 2D lines already intersect so there is no need to project them onto a plane.

For planes, we also project a 3D "anchor" point onto each plane but the anchor point is built differently: it is defined as the 3D point which coordinates are the average of the planes' offsets associated with that axis, \ie, the $x$ coordinate is the average of offsets $c$ of all planes of the form $x=c$.

\PAR{NN Learning.}
The recovery of the points hidden by obfuscated representations assumes that the original points' neighborhood information is available, \ie, one knows which obfuscations hide points that are close to each other.
The main experiments are run using an oracle that produces neighborhoods with various levels of inlier ratios to allow for the evaluation of the robustness of the recovery against inaccurate neighborhood information.
In parallel, we show that the descriptors preserved by the geometric obfuscation hold enough information to infer the neighborhood necessary for the recovery.
To do so, we train a transformer-based network to learn a similarity score between all pairs of descriptors that is inversely proportional to the distance between the original points.
A simple nearest-neighbor search using the learned similarity lets us infer nearest-neighbor points.

The network is made of 6 self-attention blocks with 4 heads.
Prior to being fed to the attention blocks, the input descriptor is first projected onto a 256-dimension space with an MLP.
The SIFT-variant of the network is trained on SIFT~\cite{Lowe04IJCV} features extracted from 97K images sampled from all the scenes of the ScanNet dataset~\cite{dai2017scannet}.
The SuperPoint-variant of the network is trained on Superpoint~\cite{Lowe04IJCV} features extracted from 309K images sampled from 184 scenes of the ScanNet dataset~\cite{dai2017scannet}.
The network is trained with a batch size of 16 for 10 epochs, with the Adam~\cite{KingBa15} optimizer with an initial learning rate of $5 \cdot 10^{-4}$ with a learning rate decay of 10\% starting the 3rd epoch and stopping once the learning rate reaches $10^{-5}$.

\PAR{Image inversion from Points.}
We used different inversion networks on the 2D and 3D structures to generate the images from the recovered points.
\textbf{In 3D}, we use the off-the-shelf inversion network provided by Pittaluga~\etal~\cite{pittaluga2019revealing}.
\textbf{In 2D}, we train a new model with the CoarseNet and RefineNet models of~\cite{pittaluga2019revealing} in conjunction.
The input to the network is a set of keypoints with associated descriptors only.
A loss function that fuses the L1 pixel loss and the LPIPS~\cite{zhang2018perceptual} perceptual loss is used, with 0.2 and 0.8 as their corresponding weights.
We train two different variants for indoor and outdoor scenes.
The indoor model is trained on 200 scenes from the ScanNet~\cite{dai2017scannet} dataset and the outdoor variant is trained on 150 scenes of Megadepth~\cite{li2018megadepth}.
Note that we do not use any of these two datasets for any evaluation.

\begin{figure*}[t]
  \centering
    \includegraphics[width=0.99\linewidth]{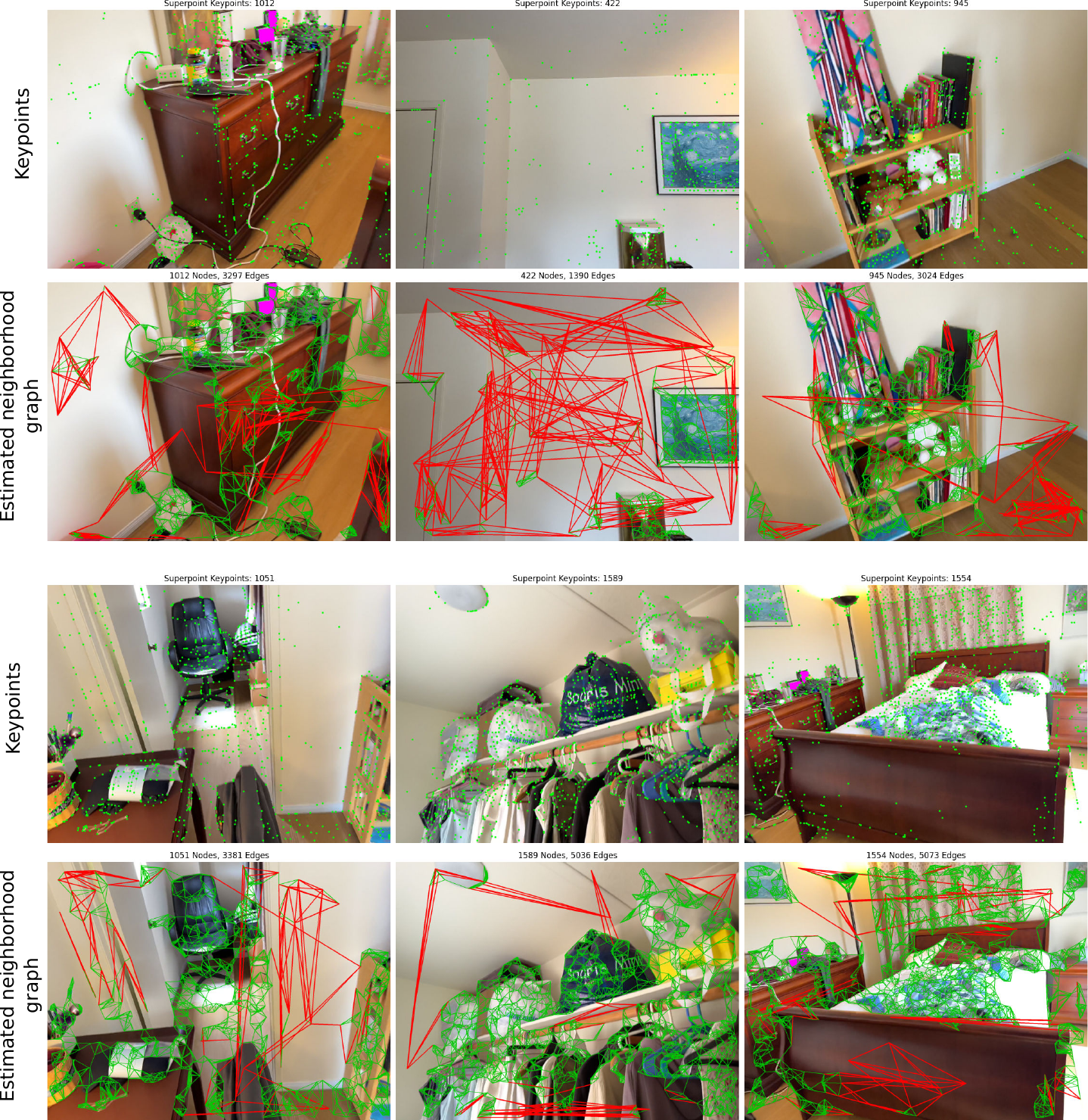}
  \caption{\textbf{Neighborhood estimated from SuperPoint~~\cite{detone2018superpoint} descriptors}: Images from scene \textit{0a76e0647} from the ScanNet++~\cite{yeshwanthliu2023scannetpp} showing detected SuperPoint~~\cite{detone2018superpoint} keypoints and the neighborhood graph estimated using our network described in Sec. 5 of the main paper. Top-5 neighbors for each point have been plotted with edges colored \textcolor{green}{green} if the points are closer than a threshold and \textcolor{red}{red} otherwise. 
  }
\label{supp:nbr_scene_bedroom}
\end{figure*}

\begin{figure*}[t]
  \centering
    \includegraphics[width=0.99\linewidth]{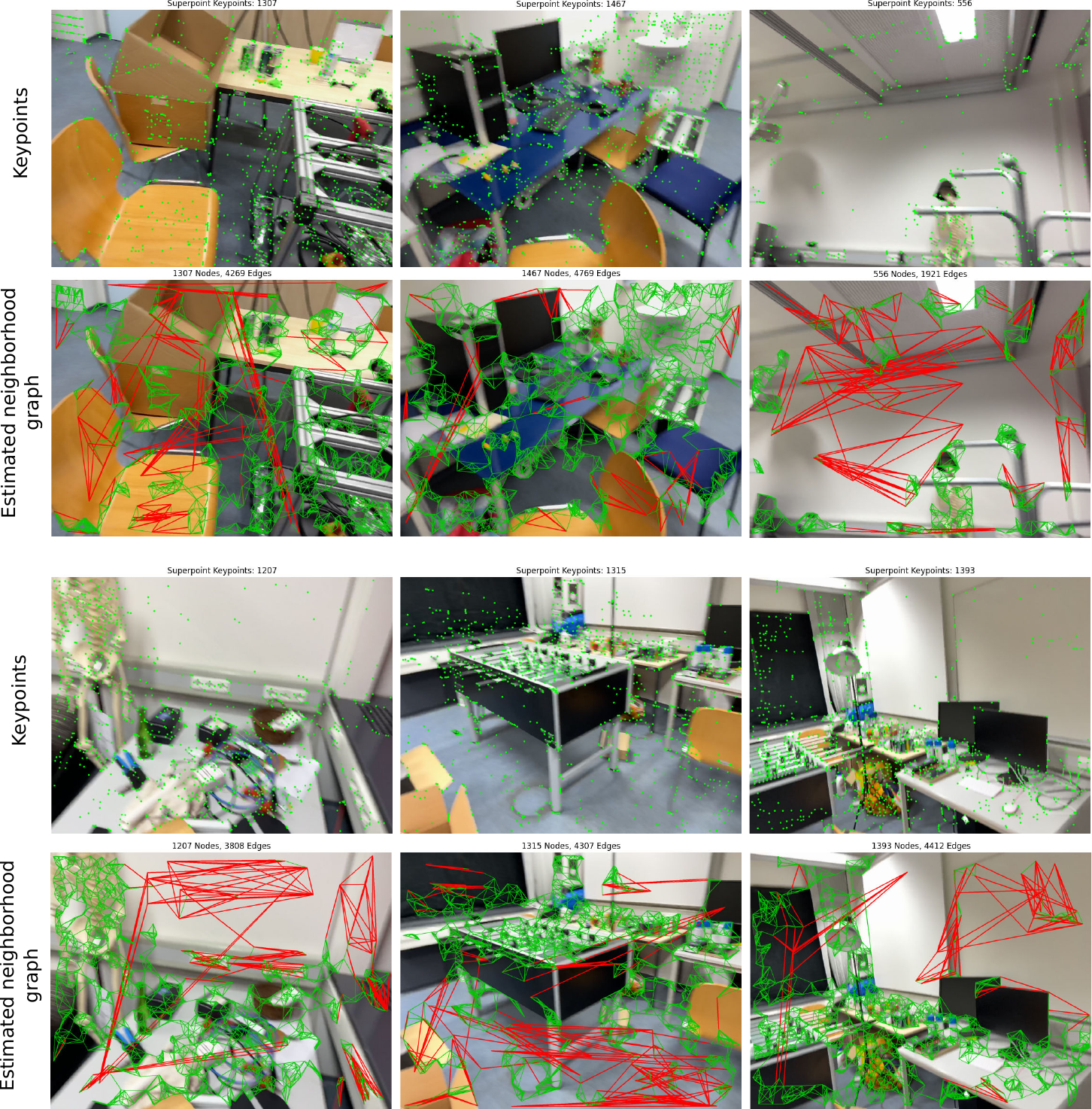}
  \caption{\textbf{Neighborhood estimated from SuperPoint~~\cite{detone2018superpoint} descriptors}: Images from scene \textit{036bce3393} from the ScanNet++~\cite{yeshwanthliu2023scannetpp} showing detected SuperPoint~~\cite{detone2018superpoint} keypoints and the neighborhood graph estimated using our network described in Sec. 5 of the main paper. Top-5 neighbors for each point have been plotted with edges colored \textcolor{green}{green} if the points are closer than a threshold and \textcolor{red}{red} otherwise. 
  }
  
\label{supp:nbr_scene_office}
\end{figure*}

\PAR{Visualizing Estimated Neighbors.}
As an additional way to  to evaluate the quality of the estimated neighborhood, we draw the neighborhood graph on top of the images in Figures~\ref{supp:nbr_scene_bedroom} and ~\ref{supp:nbr_scene_office}.
SuperPoint~\cite{detone2018superpoint} keypoints form the nodes of the graph and the graph has an edge between each point and its top-5 nearest neighbors estimated by our neighborhood estimation network (Sec. 5 of the main paper).
We use two scenes taken from the ScanNet++~\cite{yeshwanthliu2023scannetpp} dataset showing a bedroom and an office.
The accuracy of our proposed point position estimation depends on the distance of the nearest neighbors used -- the error in estimation increases if points that are far apart are considered as neighbors.
We therefore color the edges green if the distance between the corresponding nodes is less than a threshold and red otherwise.
We use $\epsilon = 0.1 * min(h,w)$ as the threshold where $h,w$ are the height and width of the image.
It is worth noting that in regions of images with more texture, such as texts, paintings, and other distinct objects, the keypoint density is high and the quality of estimated neighbors is also high.
These are regions typically containing private user content.
In texture-less parts of the scene, such as floors, walls and ceilings, the keypoints are sparse and their local regions are visually less distinct, making neighborhood estimation difficult, as illustrated by several red edges.
However, often such regions do not contain information that is private to the user.
Measures such as SSIM and PSNR treat all parts of the image equally, whereas from a privacy point of view, recovering certain parts of the image with good detail is enough to deem the method as not privacy-preserving. 
More nuanced methods to measure the privacy aspect of inverted images are therefore needed. 

\clearpage
\clearpage
{
    \small
    \bibliographystyle{ieeenat_fullname}
    \bibliography{main}
}

\end{document}